\documentclass[10pt,journal,compsoc]{IEEEtran}
\usepackage{amsmath,amsfonts,mathtools,amssymb}
\usepackage{algorithm,algorithmic}
\usepackage{array}
\usepackage{subcaption}
\usepackage{multirow}
\usepackage{textcomp}
\usepackage{stfloats}
\usepackage{url}
\usepackage{verbatim}
\usepackage[nocompress]{cite}
\usepackage{booktabs}
\usepackage{dsfont}
\usepackage{lipsum}
\usepackage[dvipsnames]{xcolor}
\usepackage{setspace}
\definecolor{blue}{HTML}{0038A8}
\usepackage[breaklinks,colorlinks,draft=False]{hyperref}
\hypersetup{
  colorlinks,
  citecolor=blue,
  linkcolor=red,
  urlcolor=magenta}

\usepackage{cleveref}
\crefname{equation}{Eq.}{Eqs.}
\DeclareMathOperator*{\argmax}{argmax}

\title{Object-Centric Differentiable Logic Networks for Abstract Visual Reasoning}
\author{
    Hanne Say Eren, Ana Ezquerro, Ozan \"{O}zdenizci 
    \thanks{
    The authors are with the Institute of Machine Learning and Neural Computation, Graz University of Technology, Graz, Austria.\protect\\
    E-mail: \{hanne.sayeren, ana.ezquerro, oezdenizci\}@tugraz.at
}}

\begin{document}

\markboth{}%
{}

\IEEEtitleabstractindextext{%
\begin{abstract}
Human visual intelligence enables efficient reasoning capabilities on abstract visual problems with bounded cognitive resources and minimal experience, and yields remarkable generalization to novel problems with its ability to draw relational analogies and modular abstractions.
In contrast, deep learning models require exhaustive task-specific training on large datasets, making them prone to exploiting visual shortcuts and resulting in poor generalization.
While neuro-symbolic approaches attempt to bridge this gap by decoupling perception from reasoning with dedicated computation units, existing solutions rely either on explicit rule supervision or parameter-heavy transformer architectures that reintroduce the computational inefficiencies they aim to solve.
We introduce \textsc{Slot-Dwn} to address these limitations, a novel neuro-symbolic architecture that unifies unsupervised object-centric perception with logic-based symbolic reasoning.
Our model utilizes a Multi-Head Differentiable Weightless Network (MH-DWN) that performs efficient reasoning entirely in Boolean space using look-up tables (LUTs).
Our framework reduces visual perception complexity and introduces a LUT-based relational bottleneck with sparse input routing to minimize computational overhead.
We evaluate \textsc{Slot-Dwn} across various abstract visual reasoning benchmark tasks targeting systematic rule generalization, sample-efficient anomaly detection, and multi-relational reasoning.
We demonstrate that \textsc{Slot-Dwn} yields competitive performance with respect to the state-of-the-art, under extreme out-of-distribution regimes, while requiring up to 52\% fewer trainable parameters, with significantly faster convergence.
\end{abstract}

}

\maketitle

\section{Introduction}

Abstract Visual Reasoning (AVR) tasks assess the ability to infer underlying rules or patterns in a given visual context and apply the acquired knowledge to novel instances~\cite{malkinski2025deep}.
These types of visual reasoning tasks are studied extensively in humans, especially via Raven's Progressive Matrices (RPMs), as a tool to measure reasoning and cognitive abilities, i.e., where one can make a \emph{meaning} out of the given context~\cite{john2003raven}.
Similarly, to measure machine intelligence, several deep learning benchmarks have been proposed such as RAVEN~\cite{zhang2019raven} and its variants~\cite{hu2021stratified,ai-raven-mesh,benny2021scale}, as well as Procedurally Generated Matrices (PGMs)~\cite{barrett2018measuring}.
Despite strong in-distribution results, which even surpass human performance in some cases~\cite{mandziuk2019deepiq}, pure deep learning approaches come with a major weakness: they learn end-to-end directly from pixels and become prone to exploiting spurious visual correlations, leading to reasoning shortcuts~\cite{geirhos2020shortcut}.
In particular, Hu \emph{et~al.}~\cite{hu2021stratified} discovered that a context-blind model that processes only the answer panels without looking at the context could achieve close to state-of-the-art accuracy, suggesting that high scores can arise from biases in the answer set rather than genuine abstract reasoning.

Existing deep learning models also require extensive training and computational resources to solve such problems, contrasting with humans, who can apply acquired knowledge to a wide range of problems with limited experience and resources~\cite{griffiths2019doing}.
Since RPMs are intended to assess abstract relational reasoning and analogy skills associated with human intelligence~\cite{hofstadter2013surfaces,gentner1983structure}, evaluating models trained extensively on large RPM datasets risks conflating generalization with task-specific learning~\cite{mitchell2021abstraction}.
Deep learning models only succeed at RPMs through exhaustive task-specific training, and mostly fail to transfer those learned skills to unseen problems.
To alleviate poor out-of-distribution (OoD) generalization, recent approaches have proposed separating the task into two sub-problems: visual perception and abstract reasoning.
However, most of them either rely on symbolic rule supervision via explicit rule annotations, object/attribute-level labels~\cite{zhang2021abstract,hersche2023neuro,xu2023abstract} or over-parameterized reasoning architectures operating in continuous space \cite{mondal2024slot,webb2023systematic,mondal2023learning}, reintroducing the resource and sample inefficiency that motivated the separation in the first place, for efficient generalization similar to humans. 

We introduce \textsc{Slot-Dwn}, an efficient neuro-symbolic architecture, to address these limitations.
Our proposed model combines a pretrained object-centric perception backbone with a Multi-Head Differentiable Weightless Network (MH-DWN) by extending DWN~\cite{bacellar2024differentiable}, in which the entire reasoning process is carried out in Boolean space using look-up tables (LUTs) only.
In the LUT-based reasoning module, we introduce a shared LUT-based relational bottleneck which makes the learned rules feature-agnostic by construction, a sparse learnable mapping which drastically reduces the parameter count of the original dense wiring~\cite{bacellar2024differentiable}, and parallel processing of learned rules via multi-head LUT layers which allow the network to abstract distinct relational patterns. 
Our contributions are summarized as follows:
\begin{itemize}
    \item We introduce \textsc{Slot-Dwn}, an object-centric neural network architecture that applies differentiable logic-based weightless networks to abstract visual reasoning problems for the first time.
    \item We demonstrate competitive performance and OoD generalization capabilities of \textsc{Slot-Dwn} with respect to state-of-the-art methods on various AVR tasks: (i) an object-level generalization task that requires abstraction and transfer of learned relations to unseen objects~\cite{webb2020emergent}, (ii) a novel odd-orientation-out (O3) task for visual anomaly detection, and (iii) a relational matrices task to assess multi-relation rule abstraction in RPM-like problems~\cite{mandziuk2019deepiq}.
    \item Our model is significantly more efficient in both training and inference in comparison to current methods for visual reasoning. We show \textsc{Slot-Dwn} to be sample efficient in learning, yielding faster convergence during training with a more lightweight architecture, and performs reasoning at inference-time entirely via sparse look-up table queries and popcounts, without floating point operations.
\end{itemize}

\section{Related Work}

\subsection{Abstract Visual Reasoning}

Early deep learning approaches treat AVR as end-to-end supervised classification, by coupling a deep neural perception backbone with a reasoning head trained to score candidate answers~\cite{barrett2018measuring,zhang2019raven}.
Because such backbones lack any relational inductive bias, the Relation Network~\cite{santoro2017simple} was integrated into WReN~\cite{barrett2018measuring}, which pairs panel embeddings and aggregates them into a single relational descriptor.
Another line of work integrates the row- and column-wise structure of the matrix directly into the architecture, either through hand-designed hierarchies \cite{zheng2019abstract,benny2021scale,hu2021stratified} or through learned gating over panel subsets~\cite{wang2020abstract}. 
In GAMR~\cite{vaishnav2022gamr}, a guided attention mechanism dynamically reasons  by shifting attention over the scene to route task-relevant information into memory, instead of encoding all context panels in a single pass.
GAMR was shown to be sample-efficient and able to demonstrate zero-shot generalization.
Orthogonal to architecture-side advancements, other methods target the training itself via techniques based on contrastive objectives~\cite{malkinski2022multi}, auxiliary rule prediction~\cite{barrett2018measuring}, data augmentation~\cite{kim2020few,malkinski2022multi}, and intrinsic prediction errors~\cite{yang2023neural} to improve perceptual grounding.

In a different line of work from discriminative methods, generative approaches can synthesize rule-consistent solutions via spatial reasoning without any auxiliary supervision~\cite{shi2026,ezquerro2026relational}, and were shown to yield strong results on AVR tasks~\cite{shi2026}.
Despite this progress, generalization still remains a challenge. 
On the PGM benchmark~\cite{barrett2018measuring}, models that approach ceiling accuracy in the in-domain regime collapse to near chance on extrapolation and the held-out attribute regimes~\cite{malkinski2025deep}, and performance degrades sharply as the training set shrinks~\cite{zhuo2022effective}.
This gap between in-distribution fitting and systematic generalization remains as one of the central problems in the research field.

\subsubsection{Object-Centric Learning and Reasoning}

Models that generalize systematically tend to share a common design, where perception and reasoning are handled separately~\cite{malkinski2025deep}.
In recent works, the first stage towards systematic perception relies on learning object-centric representations.
Rather than learning from a distributed feature map, \textit{slot attention}~\cite{locatello2020} achieves this by decomposing a scene into a small set of order-invariant \textit{slot} representations through iterative competitive attention, enabling generalization to unseen compositions when trained on object discovery and property prediction tasks.
The second stage is the relational bottleneck, where the downstream computation is restricted to relations between object embeddings rather than the embeddings themselves, so learned rules are agnostic to object identity and transfer to unseen objects.
This appears in ESBN~\cite{webb2020emergent}, which separates relational and object memory, in CoRelNet~\cite{kerg2022inductive}, which reduces reasoning to a similarity matrix, and in Abstractors~\cite{altabaa2023abstractors}, which realize it via cross-attention mechanisms.

This two-stage idea was combined in OCRA~\cite{webb2023systematic}, which computes pairwise relational embeddings from slot-based features, with position embeddings serving as an indexing mechanism, and processes them with Transformers.
Similarly for Slot-Abstractor~\cite{mondal2024slot}, the Transformer is replaced with stacked Abstractor~\cite{altabaa2023abstractors} layers to handle multiple simultaneous rules and to scale to problems with more than a hundred objects.
STSN~\cite{mondal2023learning} follows the same recipe as training a slot attention encoder and a Transformer scoring module end-to-end, without any relational bottleneck.
These models establish the state-of-the-art on AVR benchmarks.
However, all of them realize the relational bottleneck through dense attention-based computations: relations are soft inner products, reasoning is $\mathcal{O}(n^2)$ with self-attention over slots and panels, and the learned rules remain implicit in the weights.
Consequently, these models approximate discrete predicates (such as equality or progression) rather than evaluating them exactly.
Our work retains the object-centric backbone but replaces the continuous relational backend with an LUT-based architecture for efficient and sparse generalization with Boolean predicates.

\subsection{Neuro-Symbolic Reasoning}

Neuro-symbolic methods make the perception and reasoning split explicit, pairing a neural front-end with a symbolic backend to recover the \emph{systematicity} that connectionist models lack~\cite{malkinski2025deep}.
PrAE~\cite{zhang2021abstract} embeds panels into a probabilistic scene representation and executes symbolic logic over attribute distributions, but requires ground-truth rule annotations and REINFORCE to train through the non-differentiable engine.
ALANS~\cite{zhang2022alans} removes the rule annotation requirement by inducing rules through an algebraic formulation on-the-fly, restoring end-to-end differentiability while retaining a fixed scene decomposition.
NVSA~\cite{hersche2023neuro} replaces the symbolic backend with a vector-symbolic architecture, lowering computational cost.
The algebraic approach of~\cite{xu2023abstract} uses a RetinaNet detector to emit discrete attribute tuples for an invariance-based reasoner.
However, two major limitations recur with these approaches.
Firstly, the symbolic vocabulary is typically declared rather than discovered.
These systems presuppose the attribute and/or rule sets, and require corresponding annotations, which restricts them to benchmarks where such metadata exists. 
Second, the cost of the symbolic components undercuts the efficiency motivation, e.g., the detector-based approach of~\cite{xu2023abstract} exceeds 100M parameters. 
Our approach eliminates these limitations by learning predicates from data-label supervision alone, with minimal inference-time computation overhead, high training efficiency and smaller model size.

\subsection{Weightless Neural Networks}

Recent progress in logic-based neural networks has enabled end-to-end training via gradient descent.
Differentiable logic gate networks~\cite{petersen2022deep} introduced continuous-valued relaxations for logical operations to optimize logic gate neurons, while differentiable weightless networks (DWNs)~\cite{bacellar2024differentiable} alternatively train interconnected LUTs by optimizing LUT wiring structures, building upon weightless neural networks (WNNs).
WNNs consist of artificial neurons called RAM nodes, which operate on binary inputs and outputs.
Rather than computing continuous-valued weighted sums like in DNNs, RAM nodes rely on look-up tables (LUTs) to encode Boolean logic as truth tables~\cite{aleksander2009brief,susskind2022weightless}.
They combine their inputs into an address that points to a specific entry in the table, and that entry becomes the output of the node.
With $n$ inputs, a RAM node can express any of the $2^{2^n}$ possible logical functions, requiring only $2^n$ bits of memory to do so, as opposed to~\cite{petersen2022deep} which requires $2^{2^n}$ memory.

Among WNN-based architectures, DWNs~\cite{bacellar2024differentiable} stand out with their expressiveness and training efficiency.
The VC dimension of WNNs grows exponentially with the number of inputs to each LUT, which makes larger-fan-in LUTs considerably more capable learners than shallow two-input alternatives~\cite{petersen2022deep}, whose parameter cost scales double-exponentially.
To date, these models have been evaluated almost exclusively on small-scale image classification tasks, tailored to edge-deployment with latency- and energy-efficiency constraints.
Our work, for the first time, successfully applies weightless networks to AVR tasks by coupling them with an object-centric perception backbone.

\begin{figure*}
    \centering
    \includegraphics[width=\linewidth]{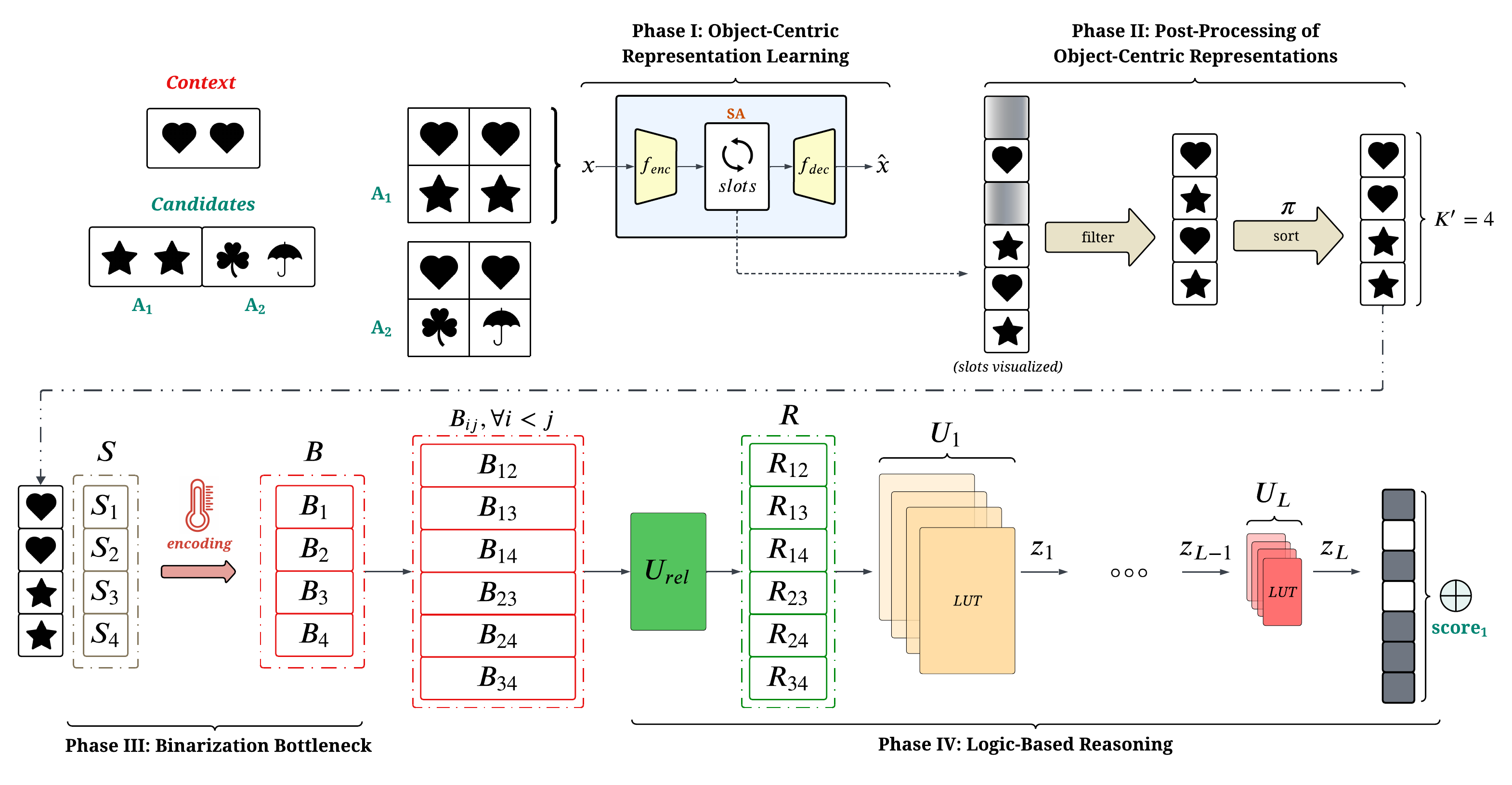}
    \caption{Our \textsc{Slot-Dwn} architecture consisting of four components. (1) An object-centric disentangled representation learning backbone that is pretrained with a reconstruction loss. (2) The pretrained \textit{slots} are post-processed to first filter out redundant slots, and then sorted by their estimated raster order for the subsequent LUT-based reasoning module. (3) A binarization bottleneck (\cref{eq:thermoenc}) converts each continuous-valued slot to a binarized representation. (4) A multi-head DWN (MH-DWN) reasoner initially processes all pairwise binary vectors in parallel with a shared LUT bottleneck $U_{rel}$ to extract relationships between objects. This relational matrix $R$ is then processed by the subsequent multi-head LUT layer(s). The score for each context and candidate pair is calculated via a final population count operation.}
    \label{fig:architecture}
\end{figure*}

\section{Methodology}

Our proposed architecture integrates unsupervised object discovery with differentiable logic-based symbolic reasoning.
The pipeline consists of four modules (see Fig.~\ref{fig:architecture}): 
(1) \textit{object-centric representation learning} via slot attention (SA), 
(2) \textit{post-processing of object-centric representations} via filtering and sorting, 
(3) a \textit{binarization bottleneck} that converts continuous-valued slots into binarized representations, and 
(4) a \textit{logic-based reasoning} module based on a Multi-Head Differentiable Weightless Neural Network (MH-DWN) architecture.

\subsection{Object-Centric Representation Learning}

We utilize Slot Attention (SA)~\cite{locatello2020} to extract object specific feature and positional information.
Initially, a convolutional encoder $f_{enc}(\cdot)$ transforms the input image $x \in \mathbb{R}^{H_x \times W_x \times C}$, where $H_x, W_x, C$ are the input's height, width and number of channels respectively, into a latent feature map $feat \in \mathbb{R}^{H \times W \times D}$.
To incorporate spatial information, a grid-based positional encoding is projected through a learnable linear layer to obtain positional embeddings $pos \in \mathbb{R}^{H\times W \times D}$.
These positional embeddings are added element-wise to the feature map.
The resulting representation is then flattened across the spatial dimensions and processed by an MLP block, to form the input to the \textit{slot attention} module, $\tilde{x} \in \mathbb{R}^{N\times D}$, where $N=(H\times W)$.

To achieve object disentanglement from the input image, the slot attention module initializes a latent bottleneck consisting of $K$ vectors, denoted as $slots \in \mathbb{R}^{K\times D}$, which are initially sampled from $\mathcal{N}(\mu,\sigma)$, where the $\mu$ and $\sigma$ are learnable.
Slots act as queries $q(slots) \in \mathbb{R}^{K\times D}$ and compete to bind to specific features of the image.
Specifically, the flattened input features $\tilde{x}$ generate keys $k(\tilde{x}) \in \mathbb{R}^{N\times D}$, and values $v(\tilde{x}) \in \mathbb{R}^{N\times D}$.
The attention weights representing the slot-to-feature mapping are calculated as:
\begin{equation}
    attn^\top = \text{softmax}\left(\frac{k(\tilde{x})\cdot q(slots)^\top}{\sqrt{D}}\right) \in \mathbb{R}^{N\times K},
    \label{eq:attn}
\end{equation}
where the softmax operates over the slot dimension, ensuring competition between slots. 
Next, slots are recurrently updated via a GRU\cite{cho2014learning} unit, using a weighted mean of $v(\tilde{x})$, where $updates=\overline{attn} \cdot v(\tilde{x})$ and $\overline{attn}$ is $attn$ normalized over the input dimension.
The slot refinement happens for $\tau$ recurrent iterations.
Finally, a spatial broadcast decoder $f_{dec}(\cdot)$~\cite{watters2019spatial} transforms these slots into individual object-specific image components and associated spatial masks $\alpha_k$, which are then recomposed into a complete, single image reconstruction $\hat{x} \in \mathbb{R}^{H_x\times W_x \times C}$.

\subsection{Post-Processing of Object-Centric Representations}
\label{subsec:postproc}

\subsubsection{Redundant Slot Filtering}
\label{subsubsec:filter}

Slot attention relies on a predefined number of slots $K$, which inherently limits its capacity to generalize to complex scenes with a varying number of objects.
This is also one of the limitations of existing object-centric learning based methods for AVR, which rely on vanilla slot attention~\cite{mondal2023learning,mondal2024slot}.
Recent studies extended slot attention to dynamically determine the optimal number of slots on a per-sample basis~\cite{fan2024adaptive,liu2026metaslot}.
Nevertheless, these approaches both introduce additional learnable parameters or masks to estimate the relevance of optimized slots, and maintain the same model complexity at inference by optimizing all $K$ slots.

We adopt an efficient post-hoc strategy for obtaining a dynamic slot count that does not require additional parameters.
After pretraining the slot-based representation learning module, we perform slot filtering based on the attention maps of slots $attn$, after $\tau$ iterative updates are done.
Specifically, for each sample, we investigate $attn_k$ indicating the spatial region of the panel where slot $k$ attends.
To filter redundant slots, we obtain the aggregate activity of each slot's attention map $attn_k \in \mathbb{R}^{N}$, and eliminate any slot with relatively small activity, e.g., $\operatorname{Avg}(attn_k) \approx 0$, concluding that the slot $k$ has attended to no meaningful region.
In practice, we apply a heuristic cut-off criterion before discarding the inactive slots.
Slots identified as redundant are removed outright rather than masked out post-decoding~\cite{fan2024adaptive}, such that the downstream model complexity scales with the true object count $K'$, instead of predefined the $K$.

\subsubsection{Sorting Slots}
\label{subsubsec:sorting}

Following the filtering step, the retained slots of an image form an unordered set, which matters for our subsequent differentiable logic-based reasoner.
As described later in Sec.~\ref{subsubsec:dwn}, our reasoning module utilizes a learnable LUT input mapping mechanism~\cite{bacellar2024differentiable}, which routes each input to a fixed coordinate that is deterministic for all samples rather than recomputed per example (\cref{eq:lm}).
Therefore, any input to the network must have a consistent meaning at each index for this routing to be learnable based on data, or the router receives arbitrary learning signals for the same address.
This holds for any input representation, not only for slots, and is what distinguishes address-based routing from transformer or attention-based models. 
Position is a natural choice for imposing this consistency, since it provides object-independent information for assigning each slot to an index.
State-of-the-art object-centric reasoning architectures treat position as an explicit signal to create a relational bottleneck independent of which object they refer to~\cite{webb2023relational}.
In~\cite{webb2023systematic,mondal2024slot}, pre-slot attention position $(pos)$ and feature $(feat)$ embeddings are factorized using the normalized attention map at the last iteration, i.e., $\widehat{feat}=\overline{attn^\tau} \cdot feat$ and $\widehat{pos}=\overline{attn^\tau} \cdot pos$. 
These factorized embeddings are then processed in separate channels for reasoning.

We adopt a similar factorization principle, where we use $\widehat{feat}$ as the object embeddings, but discard $\widehat{pos}$ and use a simple non-parametric approach instead.
Rather than integrating an explicit position embedding, we re-order the retained slots of each image into a fixed raster order and use only their feature embeddings $\widehat{feat}$ downstream.
We again utilize the attention mask of each slot for this by reshaping back to latent height and width dimensions $attn_k \in \mathbb{R}^{H\times W}$, in order to localize them by the center of mass:
\begin{equation}
    c_k = \big(c^y_k,\, c^x_k\big)
    = \frac{\sum_{i,j} attn_k[i,j]\,\big(\tilde{y}_i,\, \tilde{x}_j\big)}
           {\sum_{i,j} attn_k[i,j]}
    \;\in\; [0,1]^2 ,
    \label{eq:com}
\end{equation}
where $(\tilde{y}_i, \tilde{x}_j)$ denote the normalized spatial coordinates $\big(\frac{i}{H-1}, \frac{j}{W-1}\big)$.
Here, the $attn_k$ acts as a spatial distribution over the image and $c_k$ as the mean point where the attended object sits, independently of the object type.
The slot features $\widehat{feat}$ are then assigned a deterministic raster order according to their centroid locations: top-to-bottom, and left-to-right for slots at the same vertical position:
\begin{equation}
    k \prec l
    \iff c^y_k < c^y_l
    \;\;\lor\;\;
    \big(c^y_k = c^y_l \;\wedge\; c^x_k < c^x_l\big).
    \label{eq:raster}
\end{equation}
Let $\pi$ be the sorting function that reorders the slot features under $\prec$.
The $k$-th slot that the following reasoning module consumes corresponds to $\widehat{feat}_k = \widehat{feat}_{\pi(k)}$, where the left-hand side denotes the $k$-th object in raster order rather than the $k$-th output of the slot attention.
Thus, a given index refers to the same spatial region across every example.
We provide visual demonstrations of post-processing in Appendix~\ref{app:postproc}. 
Henceforth, we denote each continuous-valued slot feature embedding as $S_k$ instead of $\widehat{feat}_k$.

\subsection{Binarization Bottleneck}
\label{subsec:binarization}

The continuous-valued slot feature embeddings are binarized to be utilized by DWNs.
DWNs approximate gradients via the Extended Finite Difference technique~\cite{bacellar2024differentiable} (see Sec.~\ref{subsubsec:dwn}), which estimates the sensitivity of the network output to each input by flipping individual bits.
For this approximation to result in a meaningful gradient signal, it is important that nearby binary addresses correspond to semantically similar inputs, i.e., a single bit flip should represent a small, well-defined perturbation in the original embedding space rather than an arbitrary jump.

We use Thermometer Encoding~\cite{carneiro2015multilingual} to satisfy this requirement, similar to~\cite{bacellar2024differentiable}, which converts each dimension of a continuous-valued embedding $e$ into a sequence of $\beta$ indicator bits against an ordered set of thresholds $t_1 < t_2 < \cdots < t_\beta$, yielding $T(e) = (\mathds{1}_{e > t_1}, \mathds{1}_{e > t_2}, \ldots, \mathds{1}_{e > t_\beta})$.
Unlike simple thresholding, which maps each scalar to a single bit, thermometer encoding produces a unary code whose Hamming distance reflects proximity in the original embedding space: two values that differ by one threshold crossing differ in exactly one bit, while values that are far apart differ in many.
In our case, each slot feature embedding is converted dimension-wise as follows:
\begin{equation}
    T(S_k^d) = (\mathds{1}_{S_k^d > t_1^d}, \dots, \mathds{1}_{S_k^d > t_\beta^d}),  \forall d \in \{1, \dots, D\}.
    \label{eq:thermoenc}
\end{equation}
The thresholds for the thermometer encoding $T$ are determined via quantiles based on the distribution over the training data and kept fixed at inference time\footnote{As in original DWNs~\cite{bacellar2024differentiable}, we use the distributive thermometer encoding, see~\cite{bacellar2022distributive} for details.}. 
In our case, thresholds are based on learned slot feature embeddings over training data.
We denote the resulting binary representation of each $S_k$ as $B_k \in \{0,1\}^{D'}$ where $D'=D\times \beta$.
We denote the final binarized slot representation matrix as $B$ (see Fig.~\ref{fig:architecture}), composed of all $B_k$ where $k\in \{1, \dots, K'\}$.

\subsection{Logic-Based Reasoning}
\label{subsec:mh_dwn}

The binarized slot representations $B$ serve as the input to our downstream reasoning module.
We initially outline Differentiable Weightless Neural Networks (DWN)~\cite{bacellar2024differentiable}, and then extend it with a novel Multi-Head formulation (MH-DWN) to capture diverse, and hierarchically more complex interactions across the discovered object entities.

\subsubsection{Differentiable Weightless Neural Networks (DWN)}
\label{subsubsec:dwn}

Traditional WNNs rely on fixed, pseudo-random connections and are non-differentiable~\cite{aleksander2009brief}.
The recent DWN framework~\cite{bacellar2024differentiable} overcomes the scalability limitations of WNNs through two primary mechanisms, namely Learnable Mappings and Extended Finite Difference (EFD).

\textbf{Learnable Mappings:}
Dynamic routing of inputs to the LUTs in DWNs is facilitated via a learnable continuous-valued weight matrix $W \in \mathbb{R}^{P \times Q}$, where $P$ is the dimensionality of the input $z$ to the DWN layer, $Q = n \times N_U$ is the total number of input connections for $N_U$ LUTs with rank $n$, and $W_{p,q}$ is the score for routing input bit $p$ to connection $q$.
During the forward pass, the input routing $\mathcal{I}(W, z)$ is set deterministically by selecting the input bit corresponding to the maximum weight in each column $q$ of $W$:
\begin{equation}
        m_q = \argmax_{p \in \{1, \dots, P\}}W_{p, q}\qquad \mathcal{I}(W, z)_q = z_{m_q},
\label{eq:lm}
\end{equation}
During training, the upstream gradients $G$ are backpropagated to $W$ via a straight-through estimator~\cite{bengio2013estimating} considering the non-differentiable argmax, using a transformed input mapping $(2z - 1)$ which maps $\{0, 1\}$ to $\{-1, 1\}$:
\begin{equation}
    \frac{\partial \mathcal{L}}{\partial W} \approx (2z-1)^{\top} \cdot G
\end{equation}

For the backward pass of the gradient to the input $z$, softmax is applied over the first dimension $p$ of $W$:
\begin{equation}
\frac{\partial \mathcal{L}}{\partial z} \approx G \cdot \text{softmax}_p (W)^\top.
\end{equation}

\textbf{Extended Finite Difference (EFD):} 
The DWN consists of $L$ layers, each containing $N_U$ LUTs. 
We denote the $i$-th LUT of layer $l$ by $U^{(i)}_l$, where $l \in \{1, \dots, L\}$ and $i \in \{1, \dots, N_U\}$. 
Each LUT stores $2^{n}$ real-valued entries, i.e., $U^{(i)}_l \in \mathbb{R}^{2^{n}}$, and utilizes an $n$-bit binary address $a_l^{(i)}\in \{0,1\}^{n}$ to fetch a single value via an addressing function $A(U^{(i)}_l, a_l^{(i)}) : \mathbb{R}^{2^n} \times \{0,1\}^n \rightarrow \mathbb{R}$.
Since the addresses must be binary, the continuous output of each LUT is binarized by a simple $>0$ thresholding. Thus, we treat $A(\cdot, \cdot) \in \{0,1\}$ henceforth.
Backpropagation through this retrieval step is enabled by approximation of the partial derivatives via EFD~\cite{bacellar2024differentiable}.
EFD\footnote{In our experiments, we use the publicly available implementation of DWN~\cite{bacellar2024differentiable}: \href{https://github.com/alanbacellar/DWN}{https://github.com/alanbacellar/DWN}, which is based on the improved EFD calculation described here.} evaluates the derivative with respect to the address by considering the Hamming distance $H({a_l^{(i)}}', a_l^{(i)}, j)$ between the current address $a_l^{(i)}$ and all other possible memory addresses ${a_l^{(i)}}' \in \{0,1\}^{n}$, excluding the $j$-th bit:
\begin{equation}
\begin{aligned}
    \frac{\partial A\left( U^{(i)}_l,a_l^{(i)}\right)}{\partial \left(a_l^{(i)}\right)_j} = \sum_{\scalebox{0.6}{$\left({a_l^{(i)}}'\right)_j=1$}} A\left( U^{(i)}_l,{a_l^{(i)}}'\right) \cdot \kappa \eta^{H({a_l^{(i)}}', a_l^{(i)},j)} \\
    - \sum_{\scalebox{0.6}{$\left({a_l^{(i)}}'\right)_j=0$}} A\left( U^{(i)}_l,{a_l^{(i)}}'\right) \cdot \kappa \eta^{H({a_l^{(i)}}', a_l^{(i)},j)}
\end{aligned}
\label{eq:efd}
\end{equation}
and the derivative with respect to the LUT is evaluated via:
\begin{equation}
\begin{aligned}
    \frac{\partial A\left(U^{(i)}_l,a_l^{(i)}\right)}{\partial U^{(i,r)}_l} &= 
    \begin{cases} 
        1, & \text{if } r=\delta(a_l^{(i)}) \\
        0, & \text{otherwise}
    \end{cases}
\end{aligned}
\end{equation}
where $\delta$ is the function returning the LUT row index $r$, i.e., the integer equivalent of binary address $a_l^{(i)}$, and $\kappa$ and $\eta$ are exponential decay hyper-parameters\footnote{We use the default values: $\kappa=0.5 \cdot 0.75^{n-1}$, and $\eta=\frac{0.25}{0.75}$.}, tuning the sensitivity of address ${a_l^{(i)}}'$ in EFD calculation. As opposed to Finite Difference (FD) which only considers single bit flips in the address, EFD also captures address shifts to distant memory positions with an exponential decay factor.

\subsubsection{Sparse Learnable Mapping}\label{subsubsec:sparselm}

The learnable mapping $W$ at each layer requires $P\times n\times N_U$ parameters, since each LUT independently scores all $P$ candidate inputs for each of its $n$ connections.
For a fixed rank $n$, this scales when $P$ and $N_U$ increase, and makes the learnable mapping the dominant source of parameters as the network is deepened or widened.
This cost arises entirely from the size of the search space over which each connection is selected, rather than from the number of connections $n$ that an LUT actually uses.
We therefore propose to constrain the search space by restricting each LUT to a fixed \textit{routing pool} with $p\ll P$ candidate inputs, resulting in a $P/p$-fold reduction of the learnable parameters. 

This routing pool fixes which inputs an LUT can access, and therefore which relations it can express, regardless of how its mapping weights are later trained.
The key design challenge here is \emph{how} to construct the pool.
The simplest option would be letting each LUT access one contiguous block of $p$ inputs.
However, this limits an LUT to a single region and prevents it from relating inputs from different blocks.
Instead, we construct each LUT's routing pool as a \emph{strided} subsequence of the input, such that its candidates are distributed uniformly across the full input vector $P$.
We detail the strided pooling implementation in Appendix~\ref{app:sparselm}.

\subsubsection{Proposed Multi-Head DWN (MH-DWN)}

DWNs perform competitively while being highly efficient in images or simple tabular data based classification tasks~\cite{bacellar2024differentiable,soegeng2026ttsparse}.
On the other hand, object-centric reasoning requires the reasoning model to simultaneously evaluate multiple independent factors (e.g., spatial relations, feature matching) across the binary slot representations.
To allow the DWN reasoner to capture these multiple factors in parallel, we introduce a \textit{multi-head} extension of the DWN architecture.

We expand the computational capacity of each layer $l$ into $H$ independent heads.
Each head $h \in \{1, \dots, H\}$ acts as a specialized weightless sub-network for the layer-wise LUTs.
We denote the $i$-th LUT in layer $l$ and head $h$ as $U_l^{(h,i)}$.
Similarly, we follow the same notation for input dimension $P$ and output dimension $N_U$, i.e., number of LUTs.

\begin{figure}
    \centering
    \includegraphics[width=\linewidth]{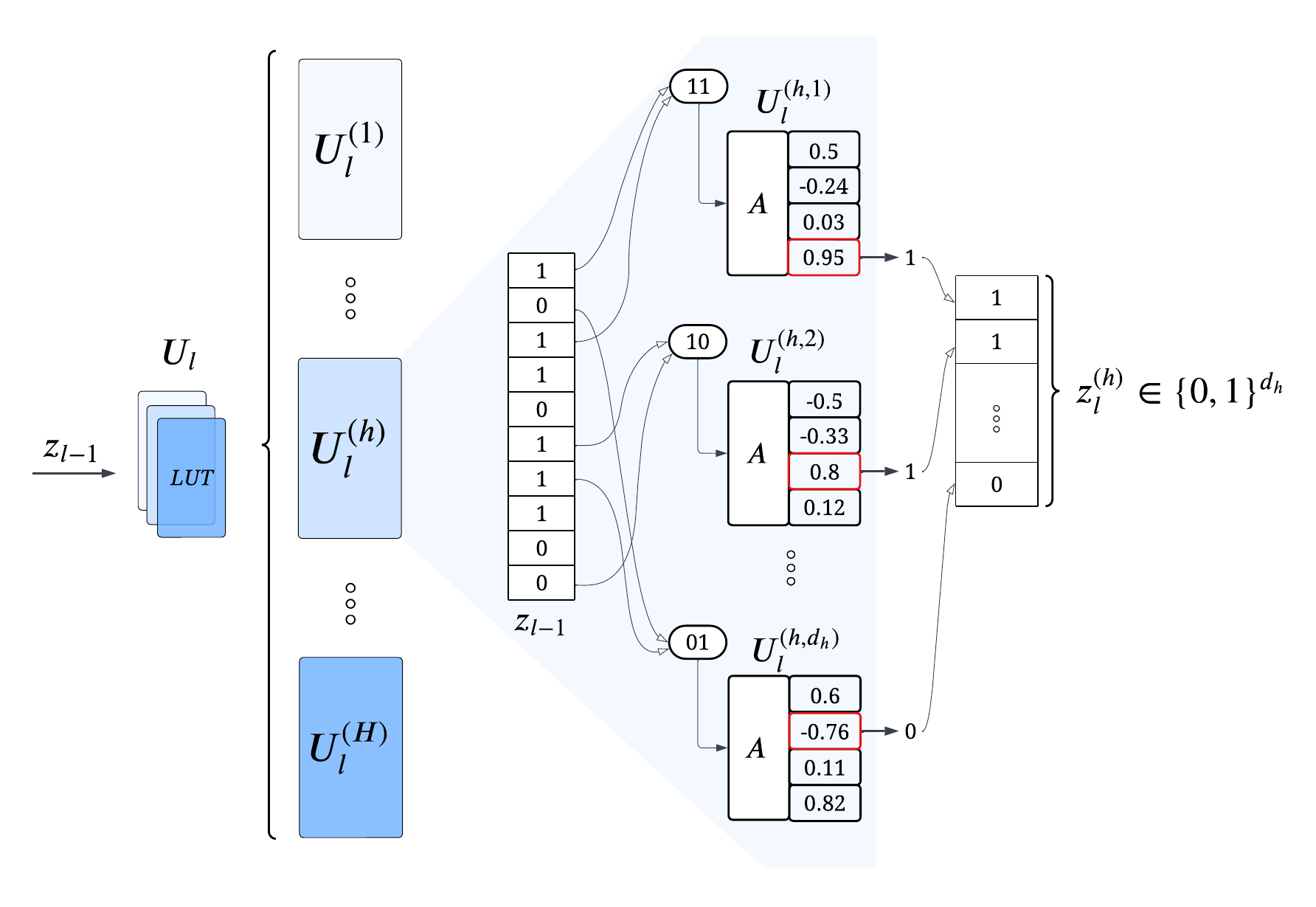}
    \caption{Illustration of a multi-head LUT layer. All LUT heads process the incoming binary vector $z_{l-1}$ individually, where the mapping wires specific input bits to each LUT $U_l^{(h,i)}$. The input bits form an address that selects a row from each LUT, yielding a binary value. All LUTs in head $h$ jointly produce $z_l^{(h)}$, and all $H$ heads form $z_l$ for the next layer.}
    \label{fig:lut}
\end{figure}

In the multi-head case, instead of a single global routing matrix, each head maintains its own learnable mapping weight matrix $W_l^{(h)} \in \mathbb{R}^{P_l \times Q_l^{(h)}}$, where the number of input connections equals $Q_l^{(h)} = n \times d_h$\footnote{We denote the number of LUTs per head as $d_h$, instead of $N_{U_l}^{(h)}$.}.
This structure allows different heads to address distinct sub-spaces of the binarized slot features $B$, and similarly the intermediate binary vectors $z$ in the following layers, enabling parallel discovery of different logical concepts.
All LUTs are evaluated based on the $n$-bit addresses $a_l^{(h,i)}$, by retrieving the stored continuous values via $A(U_l^{(h,i)}, a_l^{(h,i)})$.
Subsequently, outputs from all $H$ heads are concatenated to form the binary representation for the next layer:
\begin{equation}
    z_l = \operatorname{concat}\left(z_l^{(h=1)}, \dots, z_l^{(h=H)}\right),
\end{equation}
where $z_l^{(h)} =\operatorname{concat}\left(\{A(U_l^{(h,i)}, a_l^{(h,i)})\}_{i=1}^{d_h}\right)$. Processing of a multi-head LUT layer is visualized in Fig.~\ref{fig:lut}.
The entire multi-head structure remains end-to-end differentiable.
The EFD gradients are computed locally within the memory arrays storing $U_l^{(h)}$ of each head $h$, and the error signal backpropagates seamlessly through the concatenation and head-specific input routing matrices $W_l^{(h)}$. 

\subsubsection{Shared Pairwise Relational LUT Layer}
\label{subsubsec:shared_lut}

Visual reasoning relies primarily on the relations between objects rather than absolute object-level features or attributes.
We introduce an inductive bias to our model to harness this relational constraint architecturally.
Accordingly, the first LUT layer of our MH-DWN network is designed such that it is applied identically to each pair of binarized slot embeddings, with each LUT wired to compare corresponding bits of each pair, to extract binary information about how the two slots (objects) relate.
We create this relational bottleneck via two mechanisms.

Firstly, we perform weight sharing across pairs, by applying a single LUT layer to all slot pairs, rather than giving each pair its own learnable parameters.
Because the LUT based operator is identical for every pair, the output encoding then depends only on the two slots' content.
The second mechanism is corresponding-bit pairing.
Weight sharing forces the LUT to be relational, but does not enforce the wiring structure.
We apply a structured mapping such that, within each pair, it routes the two matching bits of the paired slot encodings to the same LUT.
Bitwise comparison of the two slots thus becomes the natural primitive of the layer.
Since each LUT receives only two bits for its addressing, they naturally become LUT-2 version of general LUT-$n$.
Mathematically, we form all unordered unique pairs of slots, yielding $M = K'(K'-1)/2$ pairs.
For the pair $(i,j)$ with $i<j$, the two binarized slots are concatenated into:
\begin{equation}
    B_{ij} = \operatorname{concat}\left(B_i, B_j\right) \in \{0,1\}^{2D'},
    \label{eq:binpair}
\end{equation}
which is then processed by the shared relational LUT layer $U_{rel}: \{0,1\}^{2D'} \to \{0,1\}^{ D'}$. 
Its parameters are shared across all pairs, such that the relational code for each pair is evaluated with the same set of learnable parameters, yielding the relational bottleneck.

The wiring of $U_{rel}$ is fixed by a structured mapping $\phi$ that assigns the input positions each LUT in $U_{rel}$ reads:
\begin{equation}
    \phi(U_{rel}^{(b)}) = \{\,b,\; b+D'\,\}, \qquad b = 1,\dots,D',
    \label{eq:struct}
\end{equation}
such that the LUT $U_{rel}^{(b)}$ receives only the $b$-th bit of $B_i$ and $B_j$. Under the concatenated layout \cref{eq:binpair}, the offset $D'$ is exactly the distance between the corresponding bits of $B_i$ and $B_j$, so $\phi$ pairs every bit position with its counterpart in the other binarized slot.
Given the structured mapping (\cref{eq:struct}), $a_{rel}^{(b)} = \{B_i^b,\, B_j^b\}$ is the address for accessing $b$-th bits.
The binary output for this bit pair is calculated via:
\begin{equation}
    R_{ij}^{(b)} = A(U_{rel}^{(b)}, a_{rel}^{(b)}),
    \label{eq:u_rel}
\end{equation}
where $R_{ij}$ denotes the binary output for $(i,j)$-th pair, and $(b)$ denotes the corresponding bit. 
Thus, the binary relational code for each pair is $R_{ij}=A(U_{rel}, a_{rel}) \in \{0,1\}^{D'}$.
Stacking each $R_{ij}$ yields the relational layer output:
\begin{equation}
    R = [R_{ij}]_{i<j} \in \{0,1\}^{M \times D'},
    \label{eq:rel_stack}
\end{equation}
which is then flattened and passed to the subsequent layers as $z_0$. 
In practice, the pairs are folded into the batch dimension so $U_{rel}$ processes all pairs of batch samples in a single pass, and its output is unfolded back into the feature dimension before the remaining layers.
Similar to other LUT layers, $U_{rel}$ can also be extended into $H$ heads where each head $h$ applies this same structured pairing but carries an independent learnable truth table, giving $H$ distinct learned pseudo-Boolean functions of every corresponding bit pair.

\begin{algorithm}[t]
\setstretch{1.1}
\caption{\textsc{Slot-Dwn}}
\label{alg:training}
\begin{algorithmic}[1]
\renewcommand{\algorithmicrequire}{\textbf{Input:}}
\renewcommand{\algorithmicensure}{\textbf{Output:}}
\renewcommand{\algorithmiccomment}[1]{\hfill\textcolor{gray}{\footnotesize$\triangleright$ #1}}
\REQUIRE input image $x \in \mathbb{R}^{H_x\times W_x\times C}$
\ENSURE output score for $x$
\item[\textbf{Define:}] slot autoencoder with $f_{enc}$ and $f_{dec}$, slot attention iterations $\tau$, number of thresholds for thermometer encoding $\beta$, LUT rank $n$, LUT heads $H$, LUT layers $L$.
\STATE $feat, pos \leftarrow f_{enc}(x)$    \COMMENT{slot encoding}
\STATE $\tilde{x} \leftarrow feat + pos$
\item [\textbf{\textit{Slot attention}}]
\STATE $slots \sim \mathcal{N}(\mu, \sigma) \in \mathbb{R}^{K \times D}$
\FOR{$t=1, \dots, \tau$}
\STATE $\left(attn^t\right)^\top \leftarrow \operatorname{softmax}_{slots}\left(\frac{1}{\sqrt{D}}k(\tilde{x}) \cdot q(slots)^\top\right)$
\STATE $\overline{attn^t} \leftarrow \operatorname{normalize}_{inputs}(attn^t)$
\STATE $updates \leftarrow \overline{attn^t} \cdot v(\tilde{x})$
\STATE $slots \leftarrow \operatorname{FF}(\operatorname{GRU}(updates, slots)) + slots$
\ENDFOR
\STATE $\hat{x}, \alpha \leftarrow f_{dec}(slots)$    \COMMENT{slot decoding}
\STATE $\widehat{feat} \leftarrow (\overline{attn^\tau}\cdot feat)$
\item [\textbf{\textit{Post-processing}}]
\STATE $\{\widehat{feat_k}\}_{k=1}^{K'} \leftarrow \operatorname{filter}(\{\widehat{feat_k}\}_{k=1}^K, \{attn_k^\tau\}_{k=1}^K)$
\STATE $\pi \leftarrow$ obtain sorting via~\cref{eq:com,eq:raster}
\STATE $\{\widehat{feat_k}\}_{k=1}^{K'} \leftarrow \left(\{\widehat{feat_k}\}_{k=1}^{K'}\right)_\pi$ \COMMENT{sort}
\STATE $\{S_k\}_{k=1}^{K'} \leftarrow \{\widehat{feat_k}\}_{k=1}^{K'}$ \COMMENT{renaming}
\item [\textbf{\textit{Binarization}}]
\STATE $B_k \leftarrow T(S_k), \quad \forall k\in\{1, \dots, K'\}$ 
\item [\textbf{\textit{Paired wiring and relations}}]
\STATE $a_{rel} \leftarrow \phi(U_{rel})$ \COMMENT{structured wiring gives addresses}
\STATE initialize $R$ and pairs $B_{ij}, \, \forall i<j$
\FORALL{$B_{ij}$}
\STATE $R_{ij} \leftarrow A(U_{rel}, a_{rel})$
\ENDFOR
\STATE $R \leftarrow [R_{ij}]_{i<j}$ \COMMENT{stacking pair relations ~\cref{eq:rel_stack}}
\STATE $z_0 \leftarrow \operatorname{flatten}(R)$
\item [\textbf{\textit{Multi-head LUT layers}}]
\FOR{$l=1, \dots, L$}
\FOR{$h=1, \dots, H$}
\STATE $z_l^{(h)} \leftarrow \operatorname{concat}\left(\{A(U_l^{(h, i)}, a_l^{(h, i)})\}_{i=1}^{d_h}\right)$
\ENDFOR
\STATE $z_l \leftarrow \operatorname{concat}\left(\{z_l^{(h)}\}_{h=1}^H\right)$
\ENDFOR
\STATE $\text{score} \leftarrow \operatorname{popcount}(z_L)$ \COMMENT{see~\cref{eq:popcount}}
\ENSURE score
\end{algorithmic}
\end{algorithm}

\subsubsection{Candidate Scoring}

Finally, a candidate scoring mechanism converts the final binary representation $z_L$ into a single scalar, which can be compared across context \& candidate pairs and used for parameter optimization within a discriminative task loss (i.e., cross-entropy over candidates).
To that end, we use the GroupSum operation, i.e., population count (popcount), similar to how DWN~\cite{bacellar2024differentiable} scores each sample for classification tasks.
The output vector $z_L$ is partitioned into $C$ equal sized groups, where $C$ is the number of classes, and the summation is performed within each group.
Afterwards, a pre-softmax temperature parameter $\lambda$ is applied as follows:
\begin{equation}
    \label{eq:popcount}
    score_c = \frac{1}{\lambda}\sum_{i\in G_c}z_i, \quad  c\in\{1, 2, \dots, C\},
\end{equation}
where $G_c$ is the $c$-th group of bits. 
For reasoning tasks with multiple candidates, where \textsc{Slot-Dwn} processes each context and candidate separately, summation is performed over the whole $z_L$ as a single group $(C=1)$ and applied to each context and candidate combination separately.

\section{Experiments}

\subsection{Datasets}

We evaluate \textsc{Slot-Dwn} on three datasets targeting different aspects of abstract visual reasoning: (i) ART~\cite{webb2020emergent}, which tests the systematic generalization capability, (ii) a novel Odd-Orientation-Out (\textsc{O3}) task which isolates a single distinguishing attribute among the samples, and (iii) a Relational Matrices (\textsc{RM}) task~\cite{mandziuk2019deepiq}, which follows the principles of RPMs to measure the reasoning ability similar to human IQ tests, on both in-domain (G-set) and out-domain instances (S-set)~\cite{matzen2010recreating} where rules over scene features are to be transferred to scenes with unseen objects. 

\subsubsection{Abstract Reasoning Tasks (ART)}

\begin{figure}[t]
    \captionsetup[subfigure]{font=scriptsize}
    \centering
    \begin{subfigure}[t]{0.48\columnwidth}
        \centering
        \includegraphics[width=0.99\columnwidth]{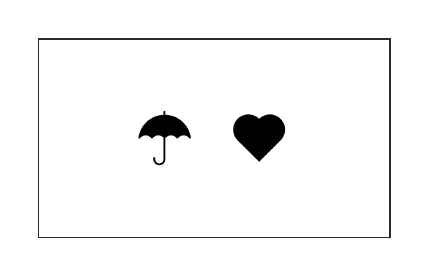}
        \caption{Same/different (\textsc{SD})}
        \label{fig:sd}
    \end{subfigure}
    \begin{subfigure}[t]{0.48\columnwidth}
        \centering
        \includegraphics[width=0.99\columnwidth]{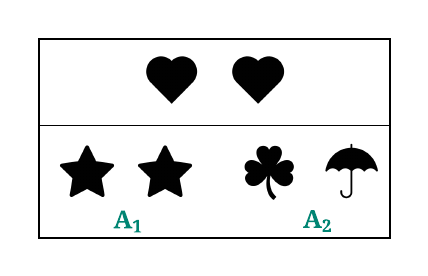}
        \caption{Relational match to sample (\textsc{RMTS})}
        \label{fig:rmts}
    \end{subfigure}\\
    \begin{subfigure}[t]{0.48\columnwidth}
        \centering
        \includegraphics[width=0.99\columnwidth]{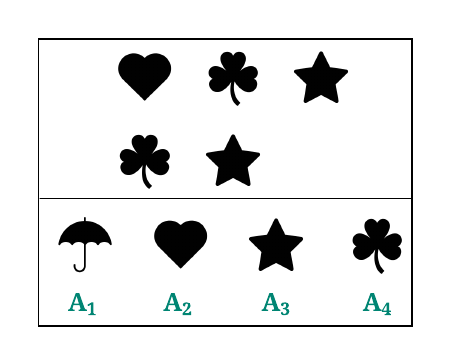}
        \caption{Distribution of 3 (\textsc{Dist-3})}
        \label{fig:dist3}
    \end{subfigure}
    \begin{subfigure}[t]{0.48\columnwidth}
        \centering
        \includegraphics[width=0.99\columnwidth]{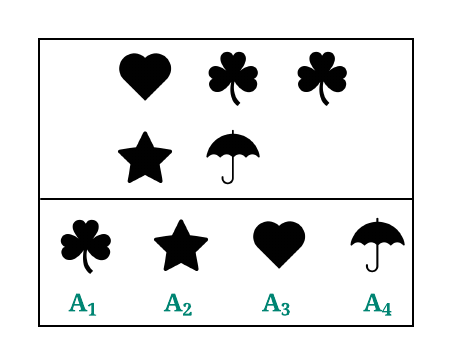}
        \caption{Identity rules (\textsc{ID})}
        \label{fig:id}
    \end{subfigure}
    \caption{Visual illustration of the \textsc{ART} tasks.}
    \label{fig:art}
\end{figure}

Abstract Reasoning Tasks (ART) dataset comprises four visual reasoning problems~\cite{webb2020emergent}, each defined by a distinct underlying rule: \textit{same/different} (\textsc{SD}), \textit{relational-match-to-sample} (\textsc{RMTS}), \textit{distribution-of-3} (\textsc{Dist-3}), and \textit{identity rules} (\textsc{ID}).
In \textsc{SD} (Fig.~\ref{fig:sd}), two objects are presented and the model must determine whether they are the same or different.
In \textsc{RMTS} (Fig.~\ref{fig:rmts}), a source pair of objects having \emph{same} or \emph{different} relationship is presented, and the model must select the target pair that shows the same relation.
In \textsc{Dist-3} (Fig.~\ref{fig:dist3}), a set of three objects appears in the first row alongside an incomplete set in the second row, the model must identify the missing object out of 3.
In \textsc{ID} (Fig.~\ref{fig:id}), the first row contains three objects following an abstract pattern (ABA, ABB, or AAA), and the model must select the object that results in the same pattern in the second row.

We evaluate the proposed model both in-distribution and out-of-distribution (OoD) cases, systematically.
For OoD evaluations, a specific number of objects are held out from the training data, denoted with $m$, allowing only the remaining objects ($100-m$) to be observed during training.
We consider $m\in\{0,95,85,50\}$ similar to previous works, where in-distribution evaluations correspond to $m=0$.

\begin{figure}[t]
    \captionsetup[subfigure]{font=scriptsize}
    \centering
    \begin{subfigure}[c]{0.48\linewidth}
        \centering
        \includegraphics[width=\linewidth]{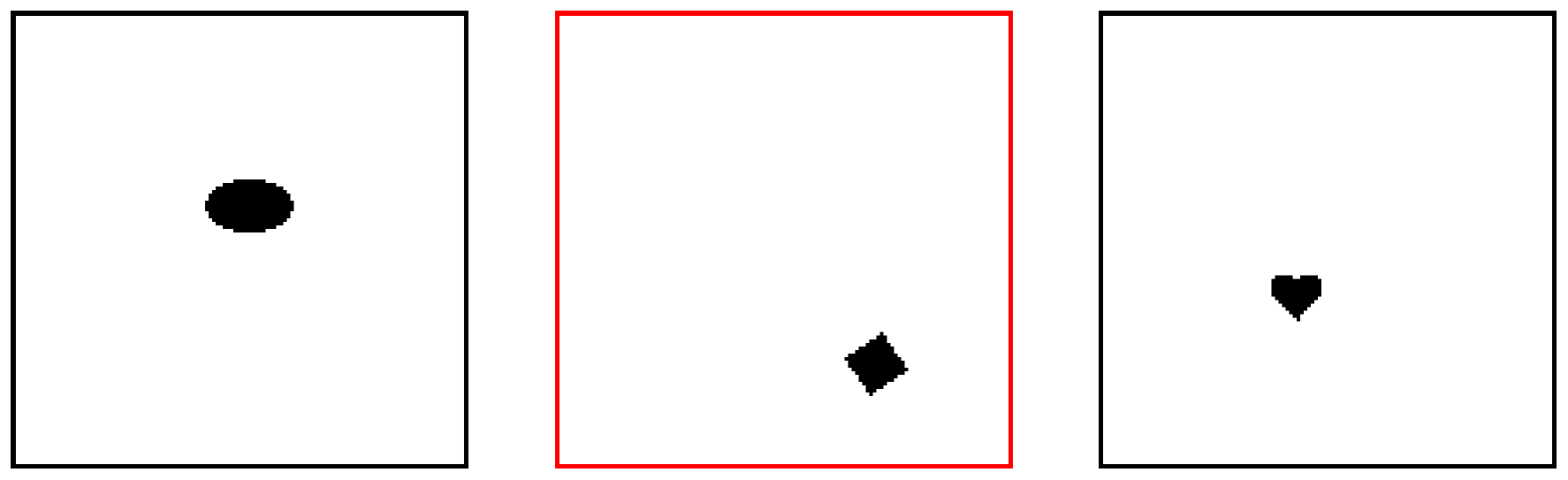}
        \caption{\textsc{O3-Easy}}
        \label{fig:o3-easy}
    \end{subfigure}\hfill
    \begin{subfigure}[c]{0.48\linewidth}
        \centering
        \includegraphics[width=\linewidth]{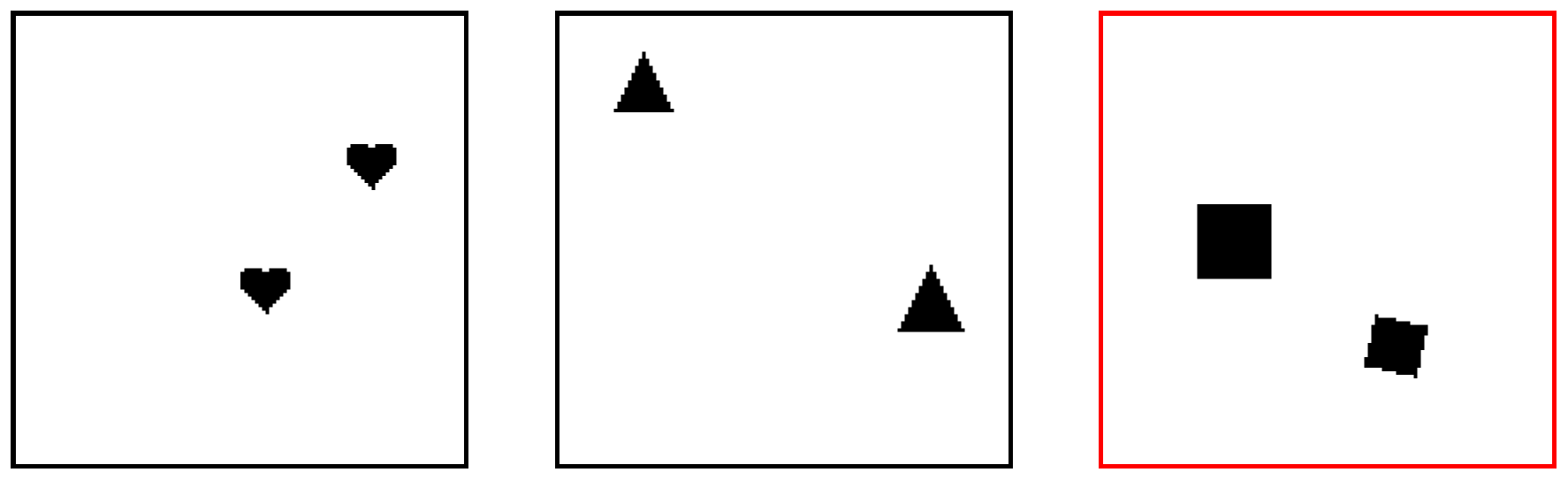}
        \caption{\textsc{O3-Hard}}
        \label{fig:o3-hard}
    \end{subfigure}
    \caption{Visual illustration of the \textsc{O3} tasks.}
    \label{fig:o3}
\end{figure}

\subsubsection{Odd-Orientation-Out (\textsc{O3})}

The \textsc{O3} task concerns identification of an odd panel among a set of three.
Unlike ART, where reasoning is defined over abstract relational rules, \textsc{O3} tasks are grounded in a single distinguishing attribute that sets one panel apart from the other two.
Objects are drawn from a set of four shapes (square, ellipse, triangle, heart), and the model is presented with three panels with the task to select the odd one.

In the \textsc{O3-Easy} task (see Fig.~\ref{fig:o3-easy}), each panel contains a single distinct object: two of the objects appear in the default orientation ($0^\circ$), while one is rotated by a random angle, and the model must identify the rotated instance.
In the \textsc{O3-Hard} task (see Fig.~\ref{fig:o3-hard}), difficulty is increased by placing two objects in each panel and it must detect the panel where only one of the objects is rotated in the odd panel.
In each \textsc{O3} task, objects vary in terms of size, position or type, so that semantic alignment rather than object identity is the only cue for the model to find the odd instance.

\begin{figure}[t]
    \centering
    \begin{subfigure}[c]{0.32\columnwidth}
        \centering
        \includegraphics[width=0.95\columnwidth]{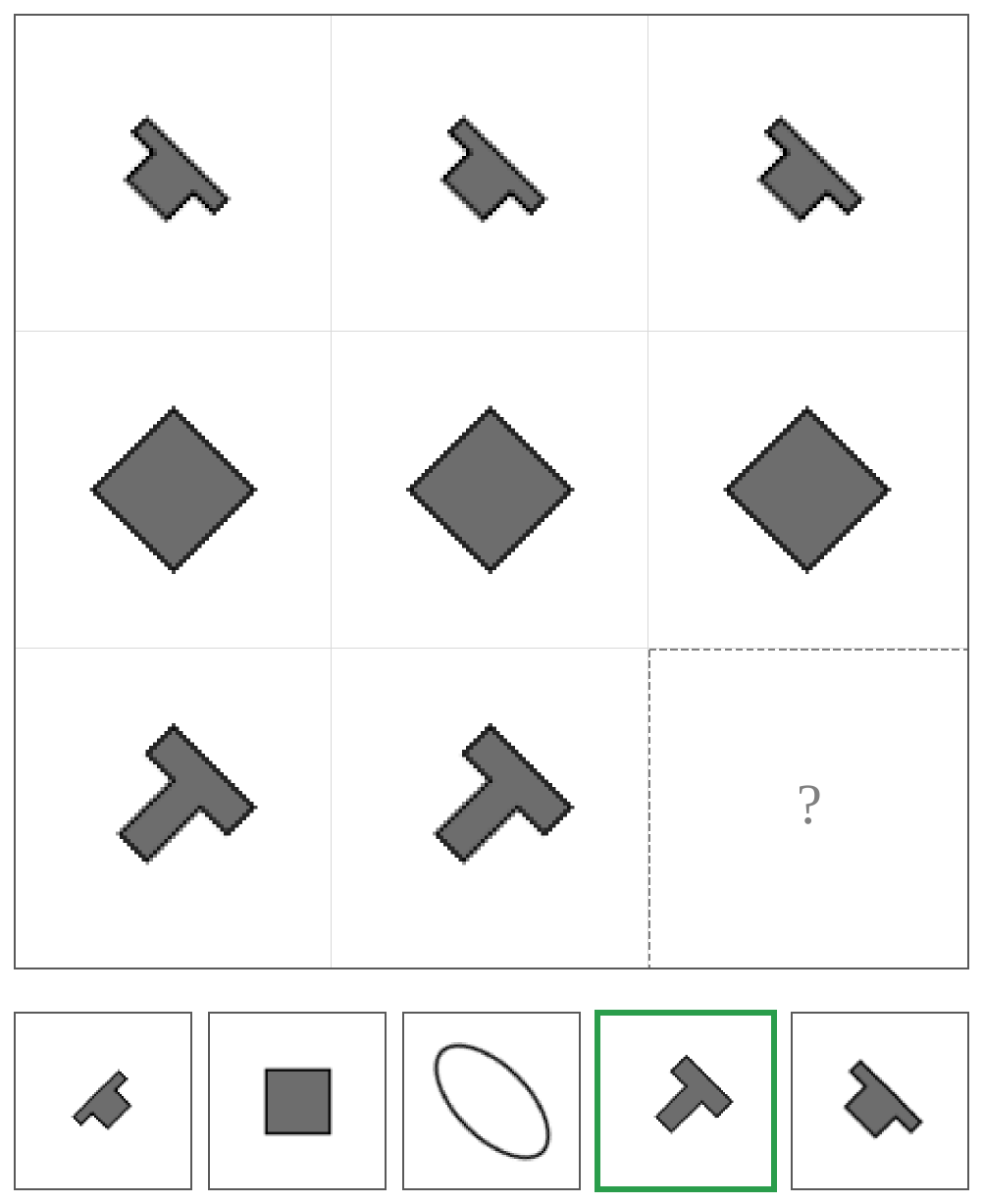}
        \caption{\textsc{RM-1}}
    \end{subfigure}
    \begin{subfigure}[c]{0.32\columnwidth}
        \centering
        \includegraphics[width=0.95\columnwidth]{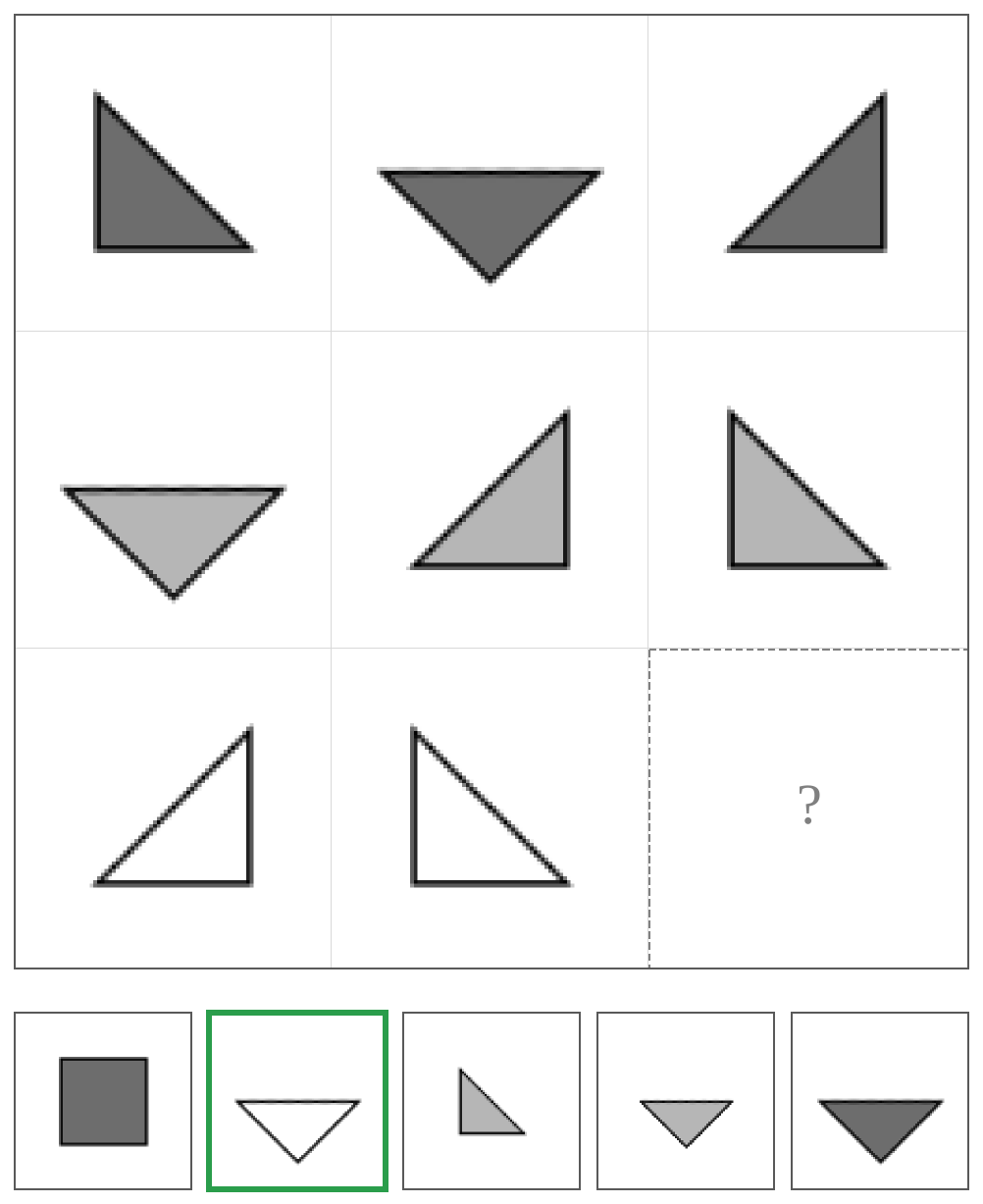}
        \caption{\textsc{RM-2}}
    \end{subfigure}
    \begin{subfigure}[c]{0.32\columnwidth}
        \centering
        \includegraphics[width=0.95\columnwidth]{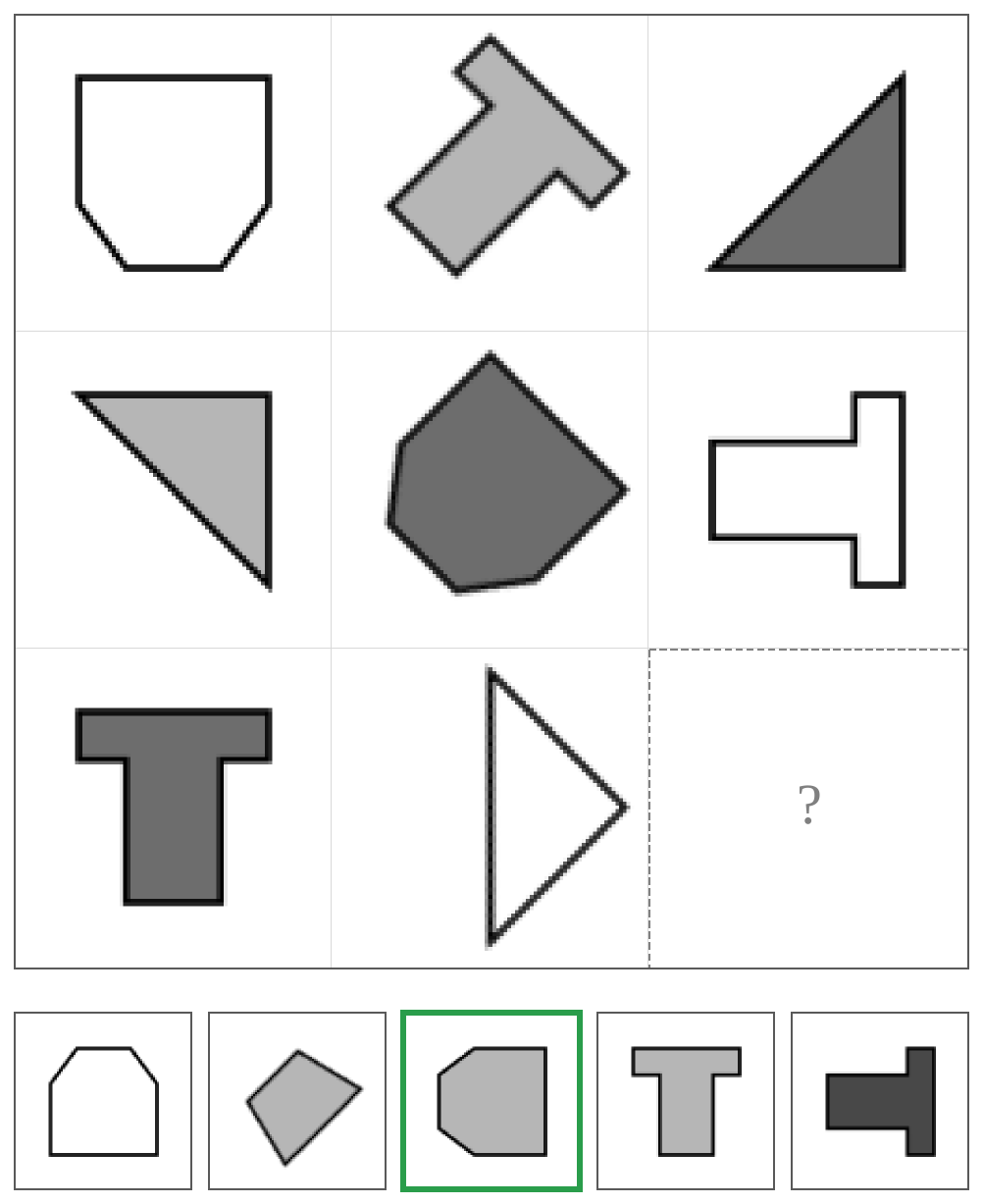}
        \caption{\textsc{RM-3}}
    \end{subfigure}
    \caption{Visual illustration of the \textsc{RM} tasks, where the number of relations between objects progressively increases.}
    \label{fig:rm-examples}
\end{figure}

\subsubsection{Relational Matrices (\textsc{RM})}

The \textsc{RM} task resembles RPMs, and adopts the G- and S-set approach of DeepIQ~\cite{mandziuk2019deepiq}. 
Each problem is a $3\times 3$ grid of panels with the bottom-right panel missing.
The model observes the context panels and must select, from a set of candidate panels, the one that completes the matrix. 
Every panel contains a single object described by four attributes: \emph{shape}, \emph{scale}, \emph{grayscale}, and \emph{rotation}. 
A problem is defined by a set of relations, each governing how each attribute changes across the matrix. 
A relation acts along one of four directions (\emph{row}, \emph{column}, \emph{left-diagonal}, or \emph{right-diagonal}), while the remaining attributes are constant along that direction.

The task difficulty is controlled by the number of relations (Fig.~\ref{fig:rm-examples}), such that the model must discover and combine several rules within a single matrix.
In these examples, attributes (scale, grayscale and rotation) are determined from a finite set of values, and shape is drawn from a fixed shape vocabulary.
The shapes are split into two disjoint groups: a \emph{training} group (G) with simple geometric shapes (e.g., rectangles, triangles), and a held-out \emph{test} group (S) of visually distinct shapes (e.g., card suits and stars) that are not observed during training (see Appendix~\ref{app:rmtrain}).
This split defines the two evaluation settings in Table~\ref{tab:rm}.
In the \textbf{in-domain setting} ($\text{G}\rightarrow\text{G}$), problems are generated from the training shapes, and the model is evaluated on a test set.
In the \textbf{out-domain setting} ($\text{G}\rightarrow\text{S}$), the model is trained similarly, but tested on problems involving held-out shapes, without fine-tuning.
Learned rules must therefore transfer to unseen objects regardless of the shape.
The attributes are shared across both groups.

\subsection{Baseline Methods}

For the ART task, similar to the baselines from~\cite{webb2023systematic}, we compare our model with both non-slot-based architectures GAMR~\cite{vaishnav2022gamr} and ResNet-50~\cite{he2016deep}, and slot-based architectures where the pretrained slot attention module serves as a backbone for varying reasoning architectures, i.e., OCRA~\cite{webb2023systematic}, Slot-GAMR~\cite{vaishnav2022gamr}, Slot-RN (Relation Network \cite{santoro2017simple}), Slot-Transformer (STSN)~\cite{mondal2023learning}, and Slot-IN (Interaction Network~\cite{watters2017visual}). 
For the \textsc{O3} and \textsc{RM} tasks, alongside OCRA and GAMR, we additionally consider Slot-Abstractors~\cite{mondal2024slot}, which differs from OCRA by combining slot attention with Abstractors~\cite{altabaa2023abstractors} as opposed to Transformer blocks. 
We also include the DeepIQ~\cite{mandziuk2019deepiq} model as a baseline in \textsc{RM} experiments, which optimizes sub-MLPs for detection of each feature separately and then measures in-domain and out-domain performance with a similar G- and S-set approach.

\begin{table*}[ht]
    \centering
    \caption{Generalization results of the baselines and \textsc{Slot-Dwn} measured at $m=95$ test set objects (i.e., only 5 objects are seen during training). Results show average and standard error $\left(\pm\right)$ of 10 runs. Best and second best values are indicated with bold and underlined text respectively. $^\dagger$ indicates STSN~\cite{mondal2023learning} with a pre-trained slot model, without end-to-end training. \# params represents the total model parameters, including the frozen perception backbone for slot-based models.}
    \label{tab:all_95}
    \resizebox{\linewidth}{!}{%
    {\renewcommand{\arraystretch}{1.1}%
    \begin{tabular}{lcrcrcrcrc}
    \toprule
     & & \multicolumn{2}{c}{\textsc{SD}} &  \multicolumn{2}{c}{\textsc{RMTS}} & \multicolumn{2}{c}{\textsc{Dist-3}} & \multicolumn{2}{c}{\textsc{ID}} \\
    \cmidrule(lr){3-4} \cmidrule(lr){5-6} \cmidrule(lr){7-8} \cmidrule(lr){9-10}
    Model & \textit{Slot} & \# params & Acc.& \# params & Acc.&  \# params & Acc. & \# params & Acc. \\
    \midrule
    \textsc{ResNet-50}\cite{he2016deep} & $\boldsymbol{\times}$ & 23.0M & 66.60 {\scalebox{0.85}{$\pm$1.5}} & 23.0M & 49.89 {\scalebox{0.85}{$\pm$0.2}} & 23.0M & 50.07 {\scalebox{0.85}{$\pm$1.3}} & 23.0M & 54.84 {\scalebox{0.85}{$\pm$2.4}}  \\
    
    \textsc{GAMR}\cite{vaishnav2022gamr} & $\boldsymbol{\times}$ & 2.70M & 83.49 {\scalebox{0.85}{$\pm$1.4}} & 2.70M & 72.20 {\scalebox{0.85}{$\pm$3.0}}  & 3.01M & 68.62 {\scalebox{0.85}{$\pm$1.8}}  & 3.01M & 66.23 {\scalebox{0.85}{$\pm$4.8}}\\
    \midrule
    \textsc{STSN}\cite{mondal2023learning, webb2023systematic}$^\dagger$ & \checkmark & 2.07M & 68.46 {\scalebox{0.85}{$\pm$2.0}}   & 2.07M & 73.99 {\scalebox{0.85}{$\pm$3.0}}   & 2.07M & 60.61 {\scalebox{0.85}{$\pm$1.9}} & 2.07M & 78.32 {\scalebox{0.85}{$\pm$1.8}}   \\
    
    \textsc{Slot-IN}\cite{watters2017visual} & \checkmark & 0.93M & 59.23 {\scalebox{0.85}{$\pm$2.3}}  & 0.93M & 56.93 {\scalebox{0.85}{$\pm$0.8}}  & 0.93M & 49.48 {\scalebox{0.85}{$\pm$1.8}}  & 0.93M & 72.82 {\scalebox{0.85}{$\pm$1.6}}   \\
    
    \textsc{Slot-RN}\cite{santoro2017simple} & \checkmark & 1.14M & 77.26 {\scalebox{0.85}{$\pm$1.9}}  & 1.14M & 61.62 {\scalebox{0.85}{$\pm$1.1}}  & 1.14M & 52.10 {\scalebox{0.85}{$\pm$0.7}}  & 1.14M & 65.96 {\scalebox{0.85}{$\pm$1.1}}   \\
    
    \textsc{Slot-GAMR}\cite{vaishnav2022gamr} & \checkmark &  2.34M & 62.98 {\scalebox{0.85}{$\pm$1.4}}  & 2.34M & 59.55 {\scalebox{0.85}{$\pm$2.7}}  & 2.35M & 32.77 {\scalebox{0.85}{$\pm$1.0}}  & 2.35M & 61.92 {\scalebox{0.85}{$\pm$0.9}}  \\
    
    \textsc{OCRA}\cite{webb2023systematic} & \checkmark & 2.07M & \underline{87.95 {\scalebox{0.85}{$\pm$1.3}}}  & 2.07M & \underline{85.31 {\scalebox{0.85}{$\pm$2.0}}}  & 2.07M & \textbf{ 86.42 {\scalebox{0.85}{$\pm$1.3}}  }& 2.07M & \textbf{92.80 {\scalebox{0.85}{$\pm$0.3}} }  \\
    \midrule
    \textsc{Slot-Dwn}  (ours) & \checkmark &  1.04M & \textbf{98.59 {\scalebox{0.85}{$\pm$0.3}} } & 1.38M & \textbf{88.49 {\scalebox{0.85}{$\pm$0.3}} } & 1.55M & \underline{84.43 {\scalebox{0.85}{$\pm$0.3}}} & 1.55M & \underline{91.45 {\scalebox{0.85}{$\pm$0.2}}} \\
    \bottomrule
    \end{tabular}}}
\end{table*}

\subsection{Training Details}
We retrieve the baseline evaluations on the ART task from OCRA~\cite{webb2023systematic}, which reports all non-slot/slot-based models, including GAMR~\cite{vaishnav2022gamr}. To evaluate the baselines on the \textsc{O3} and \textsc{RM} tasks, we keep the hyperparameters the same as in the ART task.
For Slot-Abstractor, we use the hyperparameters reported in the original work~\cite{mondal2024slot}. 
We provide details on the slot attention module pre-training and subsequent \textsc{Slot-Dwn} model training procedure below. Our code is publicly available at: \href{https://github.com/IMLTUGraz/Slot-Dwn}{https://github.com/IMLTUGraz/Slot-Dwn}.

\textbf{Slot Attention Pre-training:}
We used the pretrained slot attention model from OCRA \cite{webb2023systematic} for ART and \textsc{O3} tasks, which was trained with $K=6$ slots, $\tau=3$ iterations, $D=64$ slot dimension for auto-reconstruction with MSE. Since \textsc{O3} tasks have shapes similar to those in the ART task, we use the same pretrained slot model.
For the \textsc{RM} tasks, we trained a slot attention model with $K=11$, $\tau=3$, $D=64$. 
Slot attention model hyperparameters and optimization details for \textsc{RM} pretraining are all reported in Appendix~\ref{app:rmtrain}. 
We also include example panel and slot reconstructions of \textsc{RM} samples in Appendix Fig.~\ref{fig:slotrecons}.
After pretraining, slot model parameters remain frozen during optimization on the subsequent reasoning task.
Additionally, we observed that stochastic slot initialization~\cite{locatello2020} harmed training stability of the subsequent LUT layers, which we addressed by enforcing deterministic sampling by seeding the slot initialization.
We empirically demonstrate the importance of using deterministic slot feature embeddings in Table~\ref{tab:ablation}.

\textbf{Model Training:}
Images are resized to $128\times128$ and normalized to $[0,1]$. For ART~\cite{webb2020emergent}, following the protocol of OCRA~\cite{webb2023systematic}, samples are presented as multi-object images, where each image contains both the context and the candidate. 
This results in $2\times2$ object arrays for \textsc{RMTS} and $2\times3$ object arrays for \textsc{Dist-3} and \textsc{ID}.
We process the panels separately in \textsc{O3}.
In \textsc{RM}, we process each complete grid separately, where a candidate panel is inserted at the bottom-right alongside the context panels.

After post-processing, the true object count is reduced to $K'=2$ for \textsc{SD} and \textsc{O3-Hard}, $K'=4$ for \textsc{RMTS}, $K'=1$ for \textsc{O3-Easy}, and $K'=9$ for \textsc{RM}.
Since \textsc{Dist-3} and \textsc{ID} panels contain $K'=6$, no filtering is needed ($K=6$).
For binarization, we normalize $S$ across the slot dimension, similar to the context normalization~\cite{webb2020learning} in OCRA and GAMR.
Unlike these baselines, which add learnable affine parameters, we apply it once before training without any parameters via the fixed thermometer encoding.

All models are trained with the cross-entropy loss, using the Adam optimizer with an initial learning rate of $0.01$ and a cosine decay scheduler, with a batch size of $16$.
For \textsc{SD}, we treat it as a binary classification task, where we set popcount groups to $C=2$. 
Further hyperparameters are reported in Tables~\ref{tab:hyper} and~\ref{tab:epochsart} of the Appendix~\ref{app:hyperparams}.

\begin{table*}[t]
    \centering
     \caption{Generalization results of baselines and our model \textsc{Slot-Dwn} measured on the other generalization cases, i.e., $m=0, m=50$ and $m=85$. Our model outperforms all baselines in most cases, showing its generalization capability both in-domain and out-domain scenarios. Results show average and standard error $\left(\pm\right)$ of 10 runs. Best and second best values are indicated with bold and underlined text respectively.}
    \label{tab:overallart}
    \resizebox{\linewidth}{!}{%
    {\renewcommand{\arraystretch}{1.1}%
    \begin{tabular}{lcccccccccccc}
    \toprule
    & \multicolumn{3}{c}{\textsc{SD}} & \multicolumn{3}{c}{\textsc{RMTS}} & \multicolumn{3}{c}{\textsc{Dist-3}} & \multicolumn{3}{c}{\textsc{ID}} \\
    \cmidrule(lr){2-4} \cmidrule(lr){5-7} \cmidrule(lr){8-10} \cmidrule(lr){11-13}
    Model & $m=0$ & $m=50$ & $m=85$ & $m=0$ & $m=50$ & $m=85$ & $m=0$ & $m=50$ & $m=85$ & $m=0$ & $m=50$ & $m=85$ \\
    \midrule
    \textsc{ResNet-50}\cite{he2016deep} & 95.65 {\scalebox{0.85}{$\pm$0.7}} & 92.23 {\scalebox{0.85}{$\pm$0.6}}  & 82.83 {\scalebox{0.85}{$\pm$1.4}} &  53.77 {\scalebox{0.85}{$\pm$4.0}}  & 69.43 {\scalebox{0.85}{$\pm$6.5}}  & 82.18 {\scalebox{0.85}{$\pm$1.0}}  & 88.61 {\scalebox{0.85}{$\pm$0.3}}  & 84.32 {\scalebox{0.85}{$\pm$0.7}}  & 74.82 {\scalebox{0.85}{$\pm$1.3}}  & 93.27 {\scalebox{0.85}{$\pm$0.6}}  & 84.97 {\scalebox{0.85}{$\pm$0.6}}  & 77.94 {\scalebox{0.85}{$\pm$1.1}}  \\
    \textsc{GAMR}\cite{vaishnav2022gamr} & 97.28 {\scalebox{0.85}{$\pm$0.5}}  & 94.40 {\scalebox{0.85}{$\pm$0.3}}  & 87.88 {\scalebox{0.85}{$\pm$1.3}}  & 98.12 {\scalebox{0.85}{$\pm$0.4}}  & 91.98 {\scalebox{0.85}{$\pm$0.8}}  & 89.81 {\scalebox{0.85}{$\pm$0.5}}  & 91.98 {\scalebox{0.85}{$\pm$0.5}}  & 87.30 {\scalebox{0.85}{$\pm$0.4}}  & 80.57 {\scalebox{0.85}{$\pm$1.4}}  & 96.89 {\scalebox{0.85}{$\pm$0.5}}  & 91.90 {\scalebox{0.85}{$\pm$0.7}}  & 85.44 {\scalebox{0.85}{$\pm$1.2}}  \\
    \midrule
    
    \textsc{STSN}\cite{webb2023systematic, mondal2023learning} & \underline{98.47 {\scalebox{0.85}{$\pm$0.6}}} & 95.95 {\scalebox{0.85}{$\pm$0.5}} & 87.53 {\scalebox{0.85}{$\pm$1.1}} & \underline{99.73 {\scalebox{0.85}{$\pm$0.0}}} & \underline{99.12 {\scalebox{0.85}{$\pm$0.1}}} & \underline{96.77 {\scalebox{0.85}{$\pm$0.3}}} & \textbf{99.49 {\scalebox{0.85}{$\pm$0.0}}} & \textbf{99.08 {\scalebox{0.85}{$\pm$0.1}}} & 95.82 {\scalebox{0.85}{$\pm$0.5}} & \textbf{99.44 {\scalebox{0.85}{$\pm$0.0}}} & \textbf{99.06 {\scalebox{0.85}{$\pm$0.1}}} & \underline{97.15 {\scalebox{0.85}{$\pm$0.3}}} \\
    \textsc{Slot-IN}\cite{watters2017visual} & 94.59 {\scalebox{0.85}{$\pm$1.4}} & 92.43 {\scalebox{0.85}{$\pm$0.8}} & 68.40 {\scalebox{0.85}{$\pm$4.3}} & 93.42 {\scalebox{0.85}{$\pm$0.7}} & 87.42 {\scalebox{0.85}{$\pm$0.7}} & 86.71 {\scalebox{0.85}{$\pm$1.8}} & 97.12 {\scalebox{0.85}{$\pm$0.2}} & 96.22 {\scalebox{0.85}{$\pm$0.3}} & 90.51 {\scalebox{0.85}{$\pm$2.7}} &  97.62 {\scalebox{0.85}{$\pm$0.5}} & 96.72 {\scalebox{0.85}{$\pm$0.6}} & 94.78 {\scalebox{0.85}{$\pm$0.7}} \\
    \textsc{Slot-RN}\cite{santoro2017simple} & 93.97 {\scalebox{0.85}{$\pm$2.8}} & 94.23 {\scalebox{0.85}{$\pm$0.7}} & 83.66 {\scalebox{0.85}{$\pm$1.2}} & 91.31 {\scalebox{0.85}{$\pm$0.3}} & 87.10 {\scalebox{0.85}{$\pm$0.9}} & 68.40 {\scalebox{0.85}{$\pm$1.3}} & 97.80 {\scalebox{0.85}{$\pm$0.1}} & 96.82 {\scalebox{0.85}{$\pm$0.1}} & 91.14 {\scalebox{0.85}{$\pm$0.5}} & 93.44 {\scalebox{0.85}{$\pm$0.2}} & 91.10 {\scalebox{0.85}{$\pm$0.3}} & 86.73 {\scalebox{0.85}{$\pm$0.3}} \\
    \textsc{Slot-GAMR}\cite{vaishnav2022gamr} & 61.43 {\scalebox{0.85}{$\pm$2.7}} & 67.35 {\scalebox{0.85}{$\pm$1.1}} & 63.97 {\scalebox{0.85}{$\pm$0.5}} & 80.92 {\scalebox{0.85}{$\pm$6.9}} & 93.43 {\scalebox{0.85}{$\pm$0.4}} & 61.47 {\scalebox{0.85}{$\pm$0.8}} & 73.74 {\scalebox{0.85}{$\pm$0.7}} & 68.91 {\scalebox{0.85}{$\pm$0.6}} & 58.30 {\scalebox{0.85}{$\pm$0.5}} & 96.11 {\scalebox{0.85}{$\pm$0.1}} & 94.80 {\scalebox{0.85}{$\pm$0.2}} & 76.52 {\scalebox{0.85}{$\pm$3.0}} \\
    \textsc{OCRA}\cite{webb2023systematic} & 97.31 {\scalebox{0.85}{$\pm$1.5}} & \underline{95.98 {\scalebox{0.85}{$\pm$1.1}}} & \underline{96.48 {\scalebox{0.85}{$\pm$0.3}}} & \textbf{99.91 {\scalebox{0.85}{$\pm$0.0}}} & \textbf{99.25 {\scalebox{0.85}{$\pm$0.1}}} & 94.43 {\scalebox{0.85}{$\pm$0.7}} & 98.86 {\scalebox{0.85}{$\pm$0.2}} & 97.87 {\scalebox{0.85}{$\pm$0.3}} & \underline{96.09 {\scalebox{0.85}{$\pm$0.4}}} &  99.01 {\scalebox{0.85}{$\pm$0.0}} & 98.01 {\scalebox{0.85}{$\pm$0.1}} & 96.67 {\scalebox{0.85}{$\pm$0.2}} \\
    \midrule
    \textsc{Slot-Dwn} (ours) & \textbf{99.95 {\scalebox{0.85}{$\pm$0.0}}} & \textbf{99.80 {\scalebox{0.85}{$\pm$0.0}}} & \textbf{99.66 {\scalebox{0.85}{$\pm$0.1}}} & 98.49 {\scalebox{0.85}{$\pm$0.0}} & 99.08 {\scalebox{0.85}{$\pm$0.1}} & \textbf{97.83 {\scalebox{0.85}{$\pm$0.1}}} & \underline{99.04 {\scalebox{0.85}{$\pm$0.0}}} & \underline{98.67 {\scalebox{0.85}{$\pm$0.1}}} & \textbf{98.20 {\scalebox{0.85}{$\pm$0.1}}} & \underline{99.04 {\scalebox{0.85}{$\pm$0.0}}} & \underline{98.77 {\scalebox{0.85}{$\pm$0.1}}} & \textbf{98.18 {\scalebox{0.85}{$\pm$0.1}}} \\
    \bottomrule
    \end{tabular}}}
\end{table*}

\section{Experimental Results}

\subsection{Results on Abstract Reasoning Tasks (ART)}

Table~\ref{tab:all_95} demonstrates the results on the ART benchmark generalization evaluations, where we test the most extreme generalization ($m=95$).
Our model \textsc{Slot-Dwn} outperforms all non-slot/slot-based models in \textsc{SD} and \textsc{RMTS} tasks, and performs similarly to the state-of-the-art baseline OCRA~\cite{webb2023systematic}. 
Most importantly, our model reduces the number of trainable parameters used in OCRA at least by 26\% and further up to 52\% in \textsc{SD} case. 
Furthermore, \textsc{Slot-Dwn} outperforms OCRA and other baselines even in other generalization cases, which are depicted in Table~\ref{tab:overallart}.

We further test the generalization ability of the strongest baseline OCRA and \textsc{Slot-Dwn} in Table~\ref{tab:extreme}.
Specifically, in this extreme regime, only 2 (\textsc{SD}, $m=98$) and 4 (\textsc{Dist-3}, $m=96$) objects were observed during training.
Both models' performance drops as expected, although our model degrades far less than OCRA: on \textsc{SD} it reaches 85.33\% against 66.49\% (i.e., 13\% vs.\ 24\% relative drop from $m=95$), and on \textsc{Dist-3} it reaches 72.12\% against 65.60\% (15\% vs. 24\%). 
The performance gap becomes largest when the permutations of the context objects are removed from the training set (\textsc{Dist-3}$^\star$), where OCRA collapses to 27.44\% while \textsc{Slot-Dwn} retains 57.85\%, showing that our model depends considerably less on training data for generalization.

Finally, in Appendix~\ref{app:sanity}, we further tested genuine reasoning capabilities of \textsc{Slot-Dwn} in an ablation study, by re-defining and randomizing the correct panel labels in auxiliary tasks (i.e., $\neg$\textsc{RMTS} and \textsc{Dist-3-Rand}), which empirically supported our observations on the capability of \textsc{Slot-Dwn} to truly capture consistent relational rules.

\begin{table}[t]
    \centering
     \caption{Extreme OoD generalization results with the lowest amount of objects one can use to generate a sample. Results are averaged over 3 runs. \textsc{Dist-3}$^\star$ indicates no use of permutations of context objects in the training set.}
    \label{tab:extreme}
    \resizebox{\columnwidth}{!}{%
    {\renewcommand{\arraystretch}{1.1}%
    \begin{tabular}{lccc}
    \toprule
    Model & \textsc{SD} $\left(m=98\right)$ & \textsc{Dist-3} $\left(m=96\right)$ & \textsc{Dist-3}$^\star$ $\left(m=96\right)$\\
    \midrule
    \textsc{OCRA} & 66.49 {\scalebox{0.85}{$\pm$7.2}} & 65.60 {\scalebox{0.85}{$\pm$3.0}} & 27.44 {\scalebox{0.85}{$\pm$0.3}}\\
    \textsc{Slot-Dwn} (ours) & \textbf{85.33 {\scalebox{0.85}{$\pm$0.2}}} & \textbf{72.12 {\scalebox{0.85}{$\pm$0.4}}} & \textbf{57.85 {\scalebox{0.85}{$\pm$2.0}}}\\
    \bottomrule
    \end{tabular}}}
\end{table}

\subsubsection{Inference Efficiency}
Alongside parameter efficiency, \textsc{Slot-Dwn} is more compute-efficient at inference. 
Table~\ref{tab:flops} compares the FLOPs of slot-based reasoning models, all sharing the same frozen slot-attention front end in ART tasks. 
The reasoning stage of \textsc{Slot-Dwn} incurs almost no FLOPs across all tasks, whereas the baselines require considerably more computation. 
Since the baselines rely on dense continuous operations (matrix multiplications, attention over slots and panels), cost grows rapidly with the number of objects and embedding dimension. 
On the other hand, \textsc{Slot-Dwn} reasons entirely in Boolean space through sparse LUT look-ups on binary addresses followed by a single popcount.
The only floating-point computation is for post-processing of slots.

\subsection{Results on Odd-Orientation-Out (\textsc{O3})}\label{subsec:o3}

In \textsc{O3} experiments, instead of OoD generalization, we measure how well models perform with limited training data, while learning to distinguish the odd instance from multiple panels.
Table~\ref{tab:o3} shows the test performances of baselines GAMR~\cite{vaishnav2022gamr}, OCRA~\cite{webb2023systematic}, an additional slot-based architecture Slot-Abstractor~\cite{mondal2024slot}, and our model \textsc{Slot-Dwn}, trained under different train set sizes.
Our model outperforms all baselines, except the case \textsc{O3-Hard} (0.1K), in which the sample efficient baseline GAMR performs the best and our model follows as the second best.
OCRA and Slot-Abstractor, which are over-parameterized transformer-based architectures with up to 2.9$\times$ more parameters, suffer from limited training samples, and perform poorly as a result.
Instead, our model performs substantially better with only 100 samples for training, demonstrating that it is not only parameter efficient but also sample efficient.

\begin{table}[t]
\centering
\caption{Inference computational cost (in MFLOPs) of slot-based reasoning models. All models share the same frozen slot-attention backbone, and the rows below indicate the MFLOPs measured for reasoning only. Our model requires FLOPs only for post-processing of slots (Sec.~\ref{subsec:postproc}).}
\label{tab:flops}
 {\renewcommand{\arraystretch}{1.1}%
\begin{tabular}{lrrr}
    \toprule
    Model & \textsc{SD} & \textsc{RMTS} & \textsc{Dist-3 \& ID} \\
    \midrule
    Slot encoding & 10.8K & 21.5K & 43.0K \\
    \midrule
    \textsc{STSN}~\cite{mondal2023learning,webb2023systematic} & 17.2 & 34.4 & 68.9 \\
    \textsc{Slot-IN}~\cite{watters2017visual}   &  7.8 & 15.6 & 31.3 \\
    \textsc{Slot-RN}~\cite{santoro2017simple}   & 14.3 & 28.6 & 57.1 \\  
    \textsc{Slot-GAMR}~\cite{vaishnav2022gamr}  & 39.0 & 77.9 & 298.0 \\
    \textsc{OCRA}~\cite{webb2023systematic}  & 58.0 & 115.9 & 231.8 \\
    \midrule
    \textsc{Slot-Dwn} (ours) & \textbf{0.2} & \textbf{0.6} & \textbf{1.5} \\
    \bottomrule
\end{tabular}}
\end{table}

\begin{table*}[t]
    \centering
    \caption{\textsc{O3} task performances under varying training set sizes, shown in bold, evaluated on a fixed test set with 1K samples. All models are trained with early stopping criterion, where the patience is 10 epochs. Results show average and standard error $\left(\pm\right)$ of 3 runs. Best and second best values are indicated with bold and underlined text respectively.}
    \label{tab:o3}
     {\renewcommand{\arraystretch}{1.1}%
    \begin{tabular}{lccccc}
        \toprule
         Model & \# params & \textsc{O3-Easy} (\textbf{1K}) & \textsc{O3-Easy} (\textbf{0.1K}) & \textsc{O3-Hard} (\textbf{1K}) & \textsc{O3-Hard} (\textbf{0.1K}) \\
         \midrule
         \textsc{GAMR}\cite{vaishnav2022gamr} & 2.70M & \underline{97.59 {\scalebox{0.85}{$\pm$0.5}}} & \underline{79.23 {\scalebox{0.85}{$\pm$4.7}}} & 88.99 {\scalebox{0.85}{$\pm$7.4}} & \textbf{82.31 {\scalebox{0.85}{$\pm$3.0}}} \\
         \textsc{OCRA}\cite{webb2023systematic} & 2.07M &  90.74 {\scalebox{0.85}{$\pm$0.8}}  & 56.51 {\scalebox{0.85}{$\pm$5.9}} & \underline{93.68 {\scalebox{0.85}{$\pm$0.5}}} &  43.52 {\scalebox{0.85}{$\pm$4.7}} \\
         \textsc{Slot-Abstractor}\cite{mondal2024slot} & 2.85M & 92.49 {\scalebox{0.85}{$\pm$0.4}} & 34.82 {\scalebox{0.85}{$\pm$0.6}} & 93.61 {\scalebox{0.85}{$\pm$0.6}} & 34.03 {\scalebox{0.85}{$\pm$1.3}} \\
         \midrule
         \textsc{Slot-Dwn} (ours) & 1.04M & \textbf{97.93 {\scalebox{0.85}{$\pm$0.2}}} & \textbf{92.63 {\scalebox{0.85}{$\pm$1.3}}} & \textbf{93.90 {\scalebox{0.85}{$\pm$1.0}}} & \underline{77.45 {\scalebox{0.85}{$\pm$1.4}}}   \\
         \bottomrule
    \end{tabular}}
\end{table*}

\begin{table}[t]
    \centering
    \caption{Relational matrices (\textsc{RM}) task performance of baselines and our model \textsc{Slot-Dwn}, covering in-domain and out-domain tests (10 seeds). $^\star$FLOPs for slot encoding in object-centric methods. $^\dagger$DeepIQ~\cite{mandziuk2019deepiq} reserves feature-specific sub networks to learn the relations, whereas all other baselines use a single reasoning module overall.} 
    \label{tab:rm}
    \resizebox{\columnwidth}{!}{%
    {\renewcommand{\arraystretch}{1.1}%
    \begin{tabular}{lcccccccc}
    \toprule
     & & & \multicolumn{2}{c}{\textsc{RM-1}} &  \multicolumn{2}{c}{\textsc{RM-2}}  & \multicolumn{2}{c}{\textsc{RM-3}} \\
    \cmidrule(lr){4-5} \cmidrule(lr){6-7} \cmidrule(lr){8-9} 
    Model & \# params & MFLOPs & $\text{G} \rightarrow \text{G}$ & $\text{G} \rightarrow \text{S}$ & $\text{G} \rightarrow \text{G}$ & $\text{G} \rightarrow \text{S}$ & $\text{G} \rightarrow \text{G}$ & $\text{G} \rightarrow \text{S}$\\
    \midrule
    \textsc{DeepIQ}~\cite{mandziuk2019deepiq}$^\dagger$ & 5.66M & 73.7 & 73.30 & 70.20 & 74.10 & 71.90 & 76.00 & 73.20 \\
    \textsc{GAMR}\cite{vaishnav2022gamr} & 2.70M & 12.5K & 83.72 & 79.00 & 76.28 & 65.68 & 74.76 & 66.10 \\
    \textsc{OCRA}\cite{webb2023systematic} & 2.00M  & 13.6K$^\star$+151.3 &\underline{98.27} & \underline{87.54} & \underline{88.52} &  \underline{79.13} & \underline{84.80} & \underline{76.87}  \\
    \textsc{Slot-Abst.}\cite{mondal2024slot} & 2.78M & 13.6K$^\star$+274.4 &\textbf{99.53} & \textbf{92.27} &\textbf{92.13} & \textbf{80.02} & \textbf{88.27} & \textbf{78.33} \\
    \midrule
    \textsc{Slot-Dwn} (ours) & 1.66M & 13.6K$^\star$+0.7 & 94.42 & 83.81 & 84.72 & 75.94 & 80.27 & 70.35 \\
    \bottomrule
    \end{tabular}}}
\end{table}

\subsection{Results on Relational Matrices (\textsc{RM})}

Table~\ref{tab:rm} reports the \textsc{RM} results in the in-domain ($\text{G}\rightarrow\text{G}$) and out-domain ($\text{G}\rightarrow\text{S}$) settings for problems with one (\textsc{RM-1}), two (\textsc{RM-2}), and three (\textsc{RM-3}) relations.
All models struggle as the number of relations grows, since more independent rules must be discovered and all lose accuracy when tested on unseen shapes. 
Unlike on ART and \textsc{O3} where \textsc{Slot-Dwn} matches or outperforms the baselines in terms of generalization, in this task it outperforms GAMR~\cite{vaishnav2022gamr} and DeepIQ~\cite{mandziuk2019deepiq}, while the stronger transformer-based baselines OCRA~\cite{webb2023systematic} and Slot-Abstractor~\cite{mondal2024slot} perform best. 
The transition from $\text{G}\rightarrow\text{G}$ to $\text{G}\rightarrow\text{S}$ is similar for \textsc{Slot-Dwn} and the transformer-based baselines, which indicates that the learned rules transfer to unseen shapes about as well as them. 
In addition, the accuracy gap on \textsc{RM} should be interpreted together with the computational cost.
Our \textsc{Slot-Dwn} architecture has approximately 1M fewer parameters than Slot-Abstractor (1.66M vs. 2.78M), and its reasoning stage requires roughly $400\times$ fewer MFLOPs at inference (0.7 vs. 274.4). 
Thus, the advantage of Slot-Abstractor comes at a considerably high computational cost.

\subsection{Investigation of Symbolic Rule Learning}

We tested whether our models truly encode symbolic rules rather than only scoring candidates.
In a linear probing experiment, we trained classifier heads on the frozen latent features to predict independent properties of the rule: its direction on the grid (row, column, left- or right-diagonal) and the attribute it acts on (shape, scale, grayscale, rotation).
We used the model trained with the G-set of \textsc{RM-1} task, whose problems contain a single rule, thus, each problem had a single rule direction and attribute.
We probe: (1) the input of $U_{rel}$, i.e., flattened $B_{ij}$ pairs, (2) the output of $U_{rel}$, i.e., flattened $R$, and (3) the first MH-DWN layer output where paired relations are processed, i.e., $z_1$, by using these as the frozen latent features.
We also add two randomized controls ($R_{\mathrm{rand}}$, $z_{1,\mathrm{rand}}$) for rule-direction probing, where the LUTs of $U_{rel}$ (with weight sharing and structured wiring) and $U_1$ (with its learned wiring replaced by random wiring of the same size) are randomly initialized and kept frozen, to expose the differences in how accessible the information is.
Probes were trained on 1000 class-balanced G-set problems, and evaluated on 500 held-out G- and S-set problems.

Results in Tables~\ref{tab:probe-attribute} and~\ref{tab:probe-direction} show that relational layers primarily expose the rule direction, rather than which attribute it involves.
For the in-domain problems ($\text{G}\rightarrow\text{G}$), attributes are recoverable from $B_{ij}$ alone, especially with the problems involving scale and grayscale related rules. 
We attribute this result to the perception backbone, the slot encoder already encodes each cell's attributes, so the relational layers have little left to add about what the rule acts on. 
We also observe that shape and rotation are the weakest on the out-domain set (44.7\% and 40.1\% for $B_{ij}$). 
Nevertheless, the first LUT layer $U_1$ improves their accuracy ($z_1$: 56.5\% and 58.3\%).

Contrary to the attribute-probing results, we see that rule direction, where a relation across cells happens, is not readable from the slot pairs themselves. $B_{ij}$ reaches only 67.5\% on G-set and 64.7\% on S-set on average, whereas $R$ reaches 80.7\% and 72.4\%.
Even more, $z_1$ reaches the top performance with 86.4\% (G-set) and 80.3\% (S-set) on average.
The effect of training is clearest while transitioning from $R$ to $z_1$.
It substantially improves the probing accuracy in out-domain samples, up to 10\%.
Especially the difference between $z_1$ and its random counterpart $z_{1, \mathrm{rand}}$ is more significant, showing that learned pseudo-Boolean functions of $U_1$ help in discovering where the rule is applied.

\begin{table}[t]
\centering
\caption{Results of attribute probing (\%) on \textsc{RM-1} task, averaged over 3 seeds.}
\label{tab:probe-attribute}
 \resizebox{\columnwidth}{!}{%
{\renewcommand{\arraystretch}{1.1}%
\begin{tabular}{llcccc|c}
\toprule
Test & Model & Shape & Scale & Grayscale & Rotation & Avg. \\
\midrule
\multirow{3}{*}{$\text{G}\rightarrow\text{G}$}
 & $B_{ij}$                 & 57.0 & \textbf{97.4} & 98.0 & 64.8 & 79.3 \\
 & $R$                 & 56.1 & \textbf{97.4} & \textbf{98.4} & \textbf{67.6} & 79.9 \\
 & $z_1$                 & \textbf{65.2} & 96.6 & 96.2 & 67.4 & \textbf{81.4} \\
\midrule
\multirow{3}{*}{$\text{G}\rightarrow\text{S}$}
 & $B_{ij}$                 & 44.7 & 99.4 & \textbf{98.3} & 40.1 & 70.6 \\
 & $R$                 & 44.0 & \textbf{98.8} & 97.8 & 38.7 & 69.8 \\
 & $z_1$                 & \textbf{56.5} & 99.4 & 94.6 & \textbf{58.3} & \textbf{77.2} \\
\bottomrule
\end{tabular}}}
\end{table}

\begin{table}[t]
\centering
\caption{Results of rule direction probing (\%) on \textsc{RM-1} task, averaged over 3 seeds.}
\label{tab:probe-direction}
 \resizebox{\columnwidth}{!}{%
{\renewcommand{\arraystretch}{1.1}%
\begin{tabular}{llcccc|c}
\toprule
Test & Model & Row & Column & Left-diag. & Right-diag. & Avg. \\
\midrule
\multirow{5}{*}[-2pt]{$\text{G}\rightarrow\text{G}$}
 & $B_{ij}$                  & 74.6 & 68.1 & 60.3 & 67.0 & 67.5 \\
 & $R$                 & 89.5 & 85.1 & 69.5 & 78.8 & 80.7 \\
 & $z_1$                  & \textbf{95.4} & \textbf{86.3} & \textbf{74.6} & \textbf{89.4} & \textbf{86.4}\\
\cmidrule(lr){2-7}
 & $R_{\mathrm{rand}}$ & 83.2 & 78.7 & 65.1 & 76.4 & 75.9 \\
 & $z_{1,\mathrm{rand}}$ & 80.1 & 74.6 & 61.1 & 69.9 & 71.4 \\
\midrule
\multirow{5}{*}[-2pt]{$\text{G}\rightarrow\text{S}$}
 & $B_{ij}$                 & 67.6 & 62.8 & 64.8 & 63.5 & 64.7 \\
 & $R$                 & 78.8 & 72.2 & 68.5 & 70.1 & 72.4 \\
 & $z_1$                  & \textbf{87.6} & \textbf{82.5} & \textbf{71.3} & \textbf{79.9} & \textbf{80.3} \\
\cmidrule(lr){2-7}
 & $R_{\mathrm{rand}}$  & 73.8 & 71.5 & 68.2 & 68.2 & 70.4 \\
 & $z_{1,\mathrm{rand}}$  & 68.8 & 65.3 & 64.2 & 64.8 & 65.8 \\
\bottomrule
\end{tabular}}}
\centering
\end{table}

Overall, we can conclude that the perception backbone handles the attribute-level decoding while $U_{\mathrm{rel}}$ and $U_1$ are necessary elements for representing the rule's direction, which is needed for solving the relational problem. 

\begin{figure}[t]
    \centering
    \begin{subfigure}[t]{0.495\columnwidth}
    \centering
        \includegraphics[width=\columnwidth]{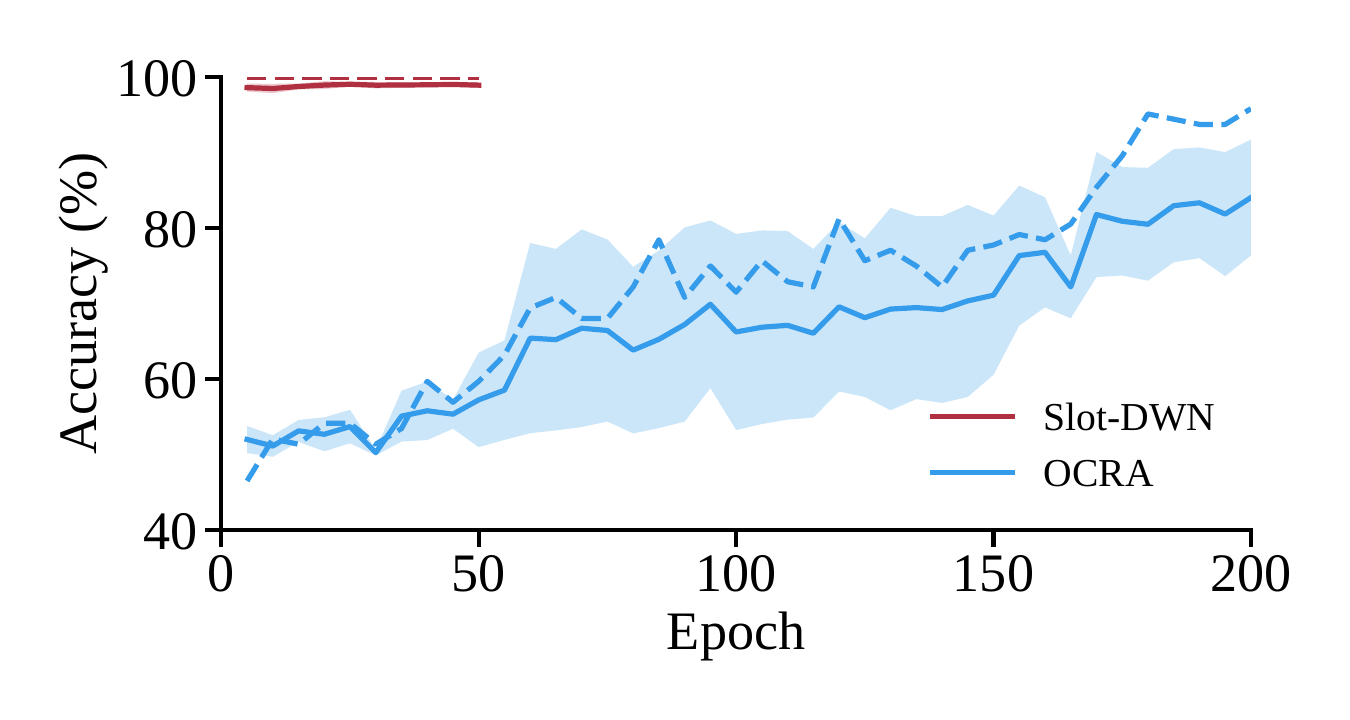}
        \caption{\textsc{SD}}
    \end{subfigure}
    \begin{subfigure}[t]{0.495\columnwidth}
    \centering
        \includegraphics[width=\columnwidth]{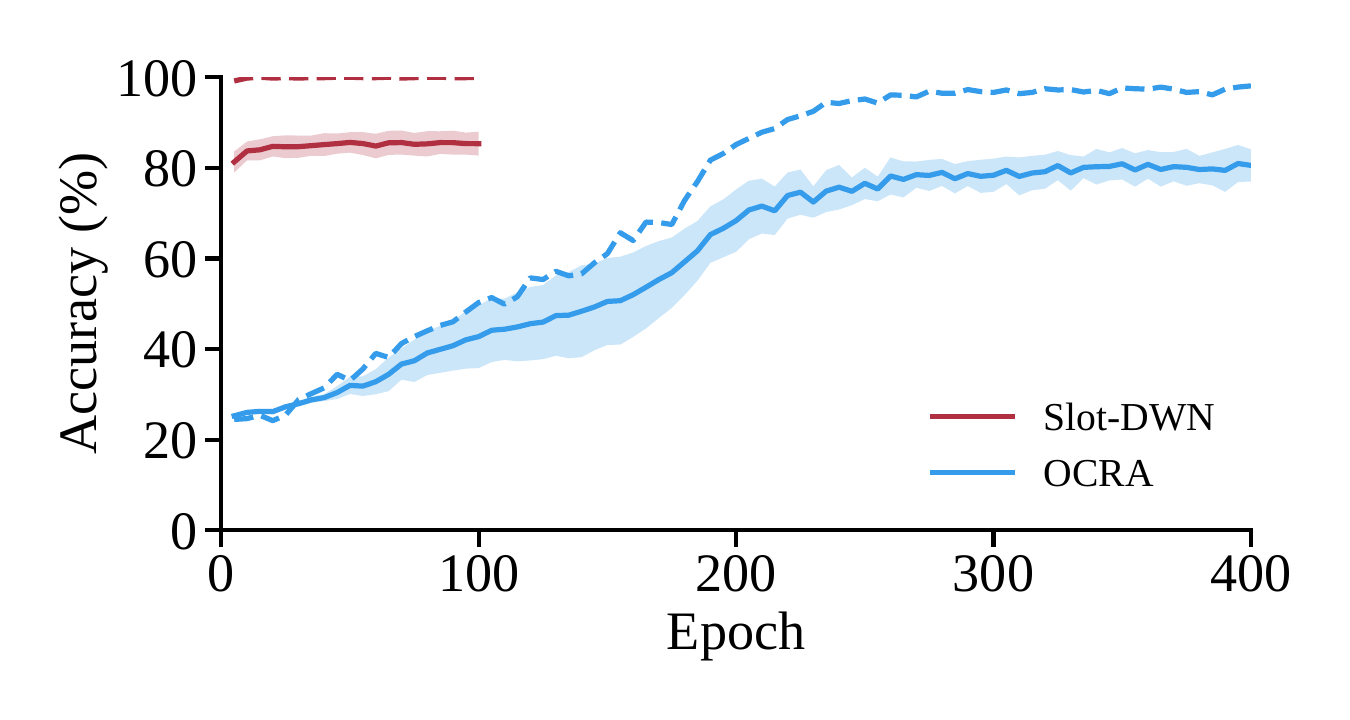}
        \caption{\textsc{Dist-3}}
    \end{subfigure}
    \vfill
    \begin{subfigure}[t]{0.495\columnwidth}
    \centering
        \includegraphics[width=\columnwidth]{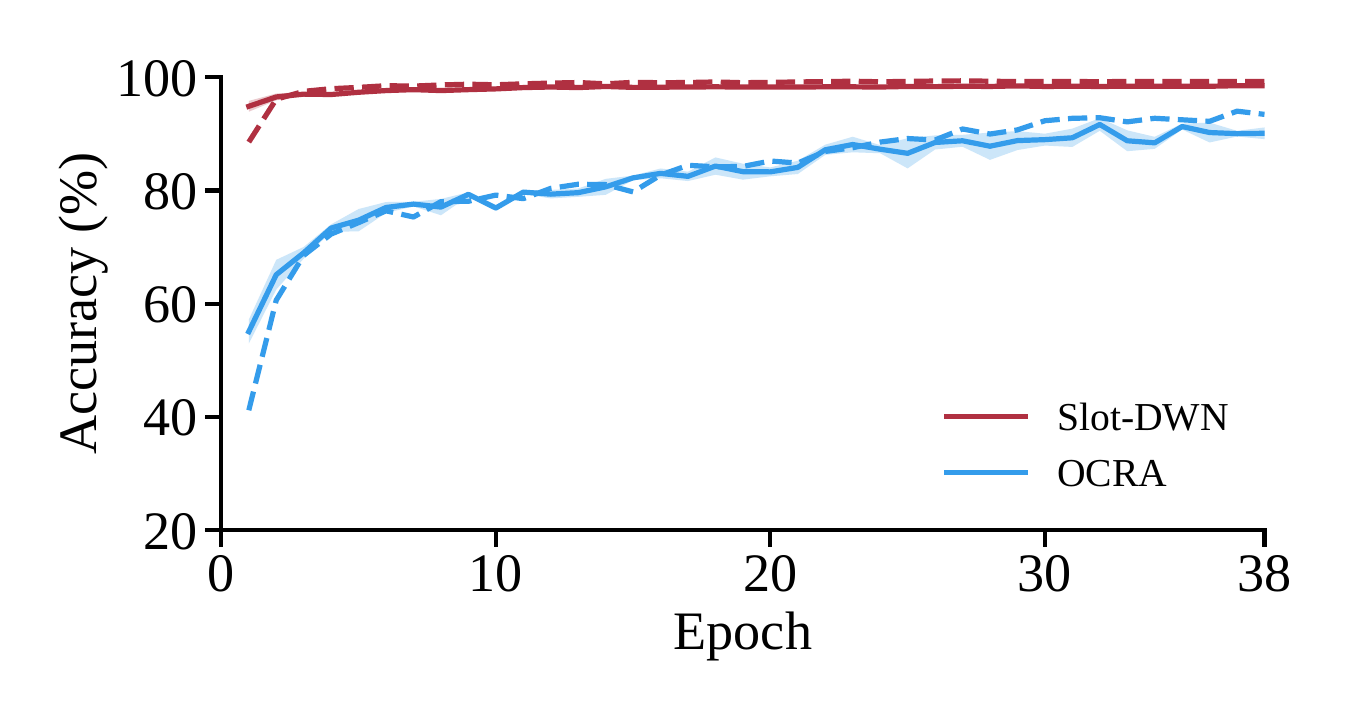}
        \caption{\textsc{O3-Easy (\textbf{1K})}}
        \label{fig:o3easy-accvsepoch}
    \end{subfigure}
    \begin{subfigure}[t]{0.495\columnwidth}
    \centering
        \includegraphics[width=\columnwidth]{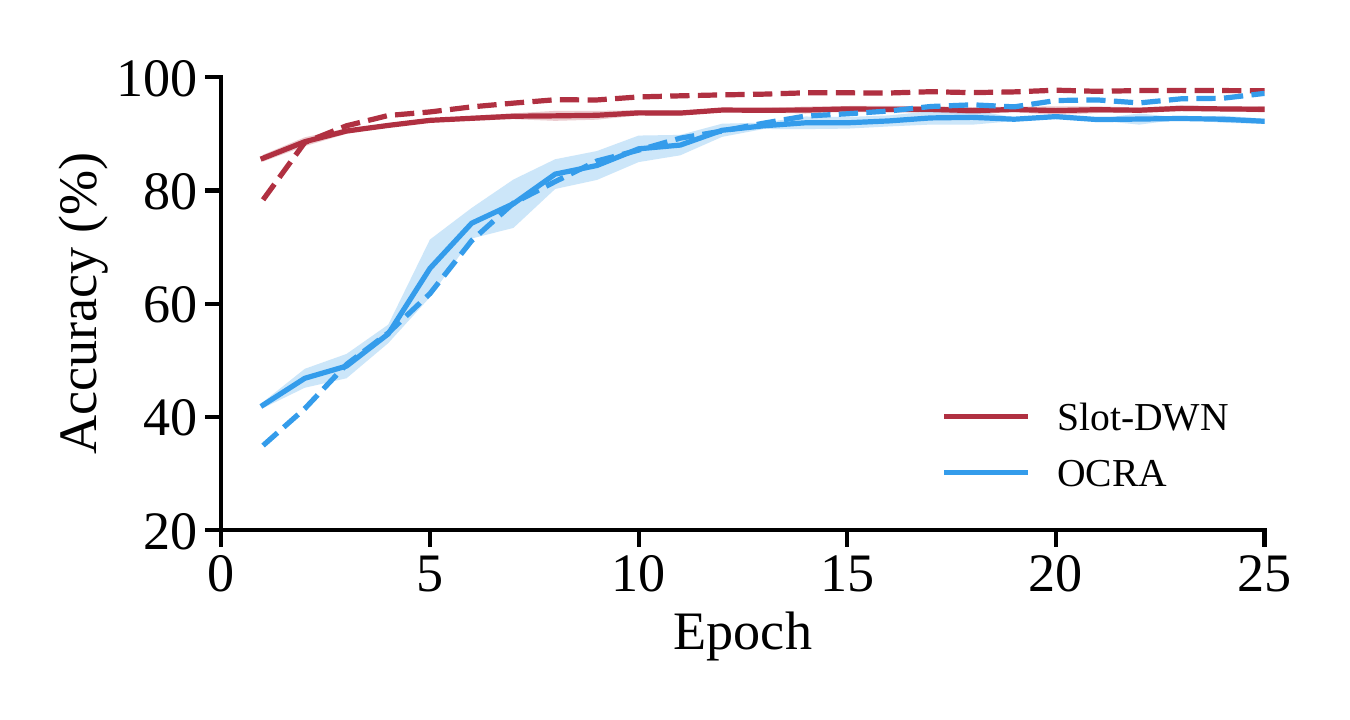}
        \caption{\textsc{O3-Hard (\textbf{1K})}}
        \label{fig:o3hard-accvsepoch}
    \end{subfigure}
    \caption{Training efficiency and convergence rate comparisons between OCRA~\cite{webb2023systematic} and \textsc{Slot-Dwn} on (a) \textsc{SD}, (b) \textsc{Dist-3}, (c) \textsc{O3-Easy} and (d) \textsc{O3-Hard}, where the models were trained with early stopping. Dashed lines indicate training and solid lines indicate test accuracy. Results are averaged over 3 runs with shaded regions indicating standard error.} 
    \label{fig:train-time}
\end{figure}

\subsection{Training Efficiency}

We evaluate training efficiency along two axes: (1) \textit{data-efficiency}, i.e., model performance under a limited amount of training data samples, and (2) \textit{convergence rate}, i.e., the number of gradient updates required for convergence.
We demonstrated the data-efficiency of our approach via our results on the \textsc{O3} dataset in Section~\ref{subsec:o3}.
When trained on a small number of samples, \textsc{Slot-Dwn} retained high accuracy and outperformed the transformer-based baselines, which have considerably more parameters yet performed poorly in the same setting.
This indicates that the model is able to extract the underlying task from limited experience.

We also examine whether this efficiency also extends to the training process itself by comparing the convergence behavior of \textsc{Slot-Dwn} with the OCRA~\cite{webb2023systematic} baseline. Fig.~\ref{fig:train-time} compares these two models on the \textsc{SD} and \textsc{Dist-3 }tasks, reporting the train and test accuracies with respect to the number of epochs.
\textsc{Slot-Dwn} reaches its final test accuracy within only 50 (\textsc{SD}) and 100 (\textsc{Dist-3}) epochs whereas OCRA needs 200 and 400 epochs respectively, i.e., 4$\times$ more updates.
Moreover, the test accuracy of OCRA varies considerably across runs, particularly on \textsc{SD}, while \textsc{Slot-Dwn} converges consistently with a narrow standard error.
We also show the convergence of both models on \textsc{O3} tasks (see Figs.~\ref{fig:o3easy-accvsepoch} and~\ref{fig:o3hard-accvsepoch}) trained with early stopping via validation set, demonstrating faster convergence of \textsc{Slot-Dwn}.

\subsection{Ablation Study}

Table~\ref{tab:ablation} reports the effect of removing individual components of \textsc{Slot-Dwn} on \textsc{Dist-3} and \textsc{RMTS}, where all variants are consistently outperformed by the full model.
In particular, we observe that object-centric perception and post-processing are critical for \textsc{Slot-Dwn}.
Replacing the object-centric backbone with a non-object-centric encoder (w/o OCL) drops accuracy 23.7\% on \textsc{Dist-3} (84.43 vs 60.78) and 31.1\% on \textsc{RMTS} (88.49 vs. 57.43), confirming that disentangled object representations are critical for reasoning generalization.
Removing the spatial sorting (w/o Sorting) causes comparable degradation (28.2\% and 26.7\%), as the learnable input mapping requires each index to refer to a consistent spatial location.
Retaining all $K$ slots (w/o Filtering) lowers \textsc{RMTS} accuracy by 16.0\% (88.49 vs 72.53), since redundant slots introduce uninformative pairs (this variant is not applicable to \textsc{Dist-3}, since $K'=K$).
Stochastic slot initialization (w/o Deterministic SA) has a relatively smaller yet substantial effect on performance, reflecting the inconsistency it induces in the binarized slot representations.

The relational components are similarly important.
Pairing each slot with itself (w/o Object Pairs) eliminates cross-slot interaction and causes a drop of over 17\% on both tasks.
Removing the structured mapping (w/o $\phi(\cdot)$), where the wiring of $U_{rel}$ is either learned or randomly initialized, results in a substantial performance drop on both tasks, showing that weight sharing alone does not provide a sufficient relational constraint.
Finally, feeding $B_{ij}$ directly to $U_1$ (w/o $U_{rel}$) decreases accuracy by 9.3\% and 15.2\%, underscoring the role of the shared relational bottleneck.
We also provide an ablation on the number of heads $H$ of MH-DWN in Appendix~\ref{app:mhdwn}, showing that an appropriate choice of $H$ for both $U_{rel}$ and the subsequent layer $U_1$ improves in-domain and out-domain performance.

\begin{table}[t]
    \centering
    \caption{Ablation studies on \textsc{RMTS} and \textsc{Dist-3}, isolating the contribution of each component of \textsc{Slot-Dwn}. Each row indicates one isolated design choice ablation. Results show average and standard error $\left(\pm\right)$ of 10 runs.}
    \label{tab:ablation}
    \scalebox{0.93}{%
    {\renewcommand{\arraystretch}{1.1}%
    \begin{tabular}{lcc}
        \toprule
          & \textsc{Dist-3} & \textsc{RMTS} \\
         \midrule
         \textsc{Slot-Dwn}& $\mathbf{84.43 \pm 0.3}$ &  $\mathbf{88.49  \pm 0.3}$ \\ 
         w/o OCL & 60.78 $\pm$ 3.4 &  57.43 $\pm$ 2.7 \\
         w/o Filtering & N/A & 72.53 $\pm$ 0.5 \\
         w/o Sorting  & 56.28 $\pm$ 2.0 & 61.79 $\pm$ 0.8 \\
         w/o Deterministic SA  & 75.48 $\pm$ 0.6 &  83.34 $\pm$ 0.7\\
         w/o Object Pairs  & 61.88 $\pm$ 2.3 & 70.88 $\pm$ 2.0 \\
         w/o $\phi(\cdot)$, learn. map. & 67.09 $\pm$ 1.3 & 65.86 $\pm$ 1.9  \\
         w/o $\phi(\cdot)$, random map. & 63.46 $\pm$ 1.4  & 60.12 $\pm$ 2.1 \\
         w/o $U_{rel}$ & 75.12 $\pm$ 2.1 &  73.30 $\pm$ 1.4 \\  
         \bottomrule
    \end{tabular}}}
\end{table}

\section{Discussion}

Our work demonstrates that systematic generalization in object-centric visual reasoning does not necessarily require the computational capacity of a dense transformer-based computation backbone, and that a powerful neuro-symbolic architecture can instead be realized with a much smaller model by exploiting discretized structures.
Since $U_{rel}$ reads only corresponding bits of two slot embeddings through a shared look-up table, its output encodes how two objects differ rather than what they are.
Even though an unseen object produces unfamiliar bit patterns, they will still end up with familiar agreement patterns (e.g., same or not same), which explains the small degradation as the number of training objects shrinks to the extreme and the robustness when context permutations are withheld. 
The transformer-based baselines, in contrast, require rich training data to learn the underlying rules (see Tables~\ref{tab:extreme} and~\ref{tab:o3}).

Probing and ablation experiments show that the structured wiring $\phi(\cdot)$ is one of the key inductive biases behind this generalization.
Each LUT in $U_{rel}$ takes two inputs where $\phi(\cdot)$ connects them to the same bit position in both object embeddings.
Regardless of what a LUT in $U_{rel}$ contains, its output is a function of a single feature evaluated on both objects.
This restricts $U_{rel}$ to comparing corresponding features, so the relational structure is determined by the wiring rather than by the table contents itself.
As observed in the rule-direction probing control cases (see Table~\ref{tab:probe-direction}), replacing the learned Boolean functions of $U_{rel}$ with random outputs $(R_\mathrm{rand})$ shows performance similar to that of $R$, the output of the learned functions.
Moreover, replacing $\phi(\cdot)$ with learned or random wiring causes a significant performance drop, even when the shared $U_{rel}$ layer is retained (see Table~\ref{tab:ablation}). 
However, this pairwise structured wiring could cause a bottleneck, as in the case of \textsc{RM}.
Our model degrades as much as the transformer-based baselines on unseen shapes, but starts from a lower in-domain accuracy and the gap widens with the number of relations, even though the learned rules transfer to out-domain shapes.
We observe that \textsc{Slot-Dwn} has relatively limited ability to express and combine multiple relations at the same time.

\subsection{Limitations}

On larger scale problems such as \textsc{RM}, \textsc{Slot-Dwn} did not consistently yield performance superior to more compute-heavy approaches, such as OCRA~\cite{webb2023systematic} and Slot-Abstractor~\cite{mondal2024slot}.
We hypothesize that a pairwise layer ($U_{rel}$) comparing only corresponding bits limits the composition of rules acting on different attributes and directions.
Higher-order relational layers are therefore a natural next step. 
Moreover, $U_{rel}$ evaluates all unordered slot pairs.
Although each LUT operation is sparse, the compute cost still grows quadratically with the object count and limits scalability, which motivates future work on learnable pair pruning.

Our pretrained slot attention module remains frozen during reasoning and the thermometer thresholds are fixed from the training distribution.
Without end-to-end task-specific gradients or finetuning, the reasoner strictly depends on the quality of the unsupervised object disentanglement.
Our ablation on stochastic slot initialization shows that the reasoner is also sensitive to inconsistencies of the binarized slots.
We argue that joint finetuning of the backbone through a straight-through binarization path, or learning the thresholds, might reduce this dependence.
Additionally, although post-hoc filtering adapts the downstream complexity to the true object count $K'$, the slot-based perception module is still bounded by a fixed choice of the maximum slot budget $K$, and \textsc{Slot-Dwn} is fixed to $K'$ after training.
The framework therefore cannot natively handle scenes with a varying number of objects, which adaptive slot mechanisms~\cite{fan2024adaptive,liu2026metaslot} might address.

Finally, the learnable mappings require a consistent slot order, which we obtain from a deterministic raster ordering of attention centroids.
This heuristic suits the grid-like layouts of our AVR tasks, but it may not preserve semantic consistency in scenes with overlapping or irregularly arranged objects.
Thus, extensions to other RPM benchmarks~\cite{zhang2019raven,hu2021stratified,barrett2018measuring} and real-world AVR tasks, remain open for future work.

\section{Conclusion}

We introduced \textsc{Slot-Dwn}, a neuro-symbolic architecture that for the first time bridges unsupervised object-centric perception with a novel Multi-Head Differentiable Weightless Network (MH-DWN), which reasons entirely in Boolean space. A shared LUT layer that processes binary slot pairs acts as a relational bottleneck, and sparse routing keeps the reasoner small. On systematic generalization and sample-efficient odd-one-out detection, the model is on par with or better than transformer-based slot reasoners, with up to 52\% fewer trainable parameters, faster convergence, and reasoning executed free of matrix multiplication and attention. On the more challenging multi-relational problems, it shows competitive results and abstraction of rules, where it transfers to out-domain contexts. These findings indicate that a relational inductive bias encoded in discrete structure can replace much of the capacity of dense attention. Future work will extend the LUT based reasoner to higher-order relations and train perception and reasoning end-to-end.

\section*{Acknowledgments}

This research was funded in whole by the Austrian Science Fund (FWF) [10.55776/COE12], and supported by the ELLIS Unit Graz and the Graz Center for Machine Learning (GraML).
The computational results have been achieved using the Austrian Scientific Computing (ASC) infrastructure.

\bibliography{refs}
\bibliographystyle{IEEEtran}

\begin{IEEEbiography}[{\includegraphics[width=1in,height=1.25in,clip,keepaspectratio]{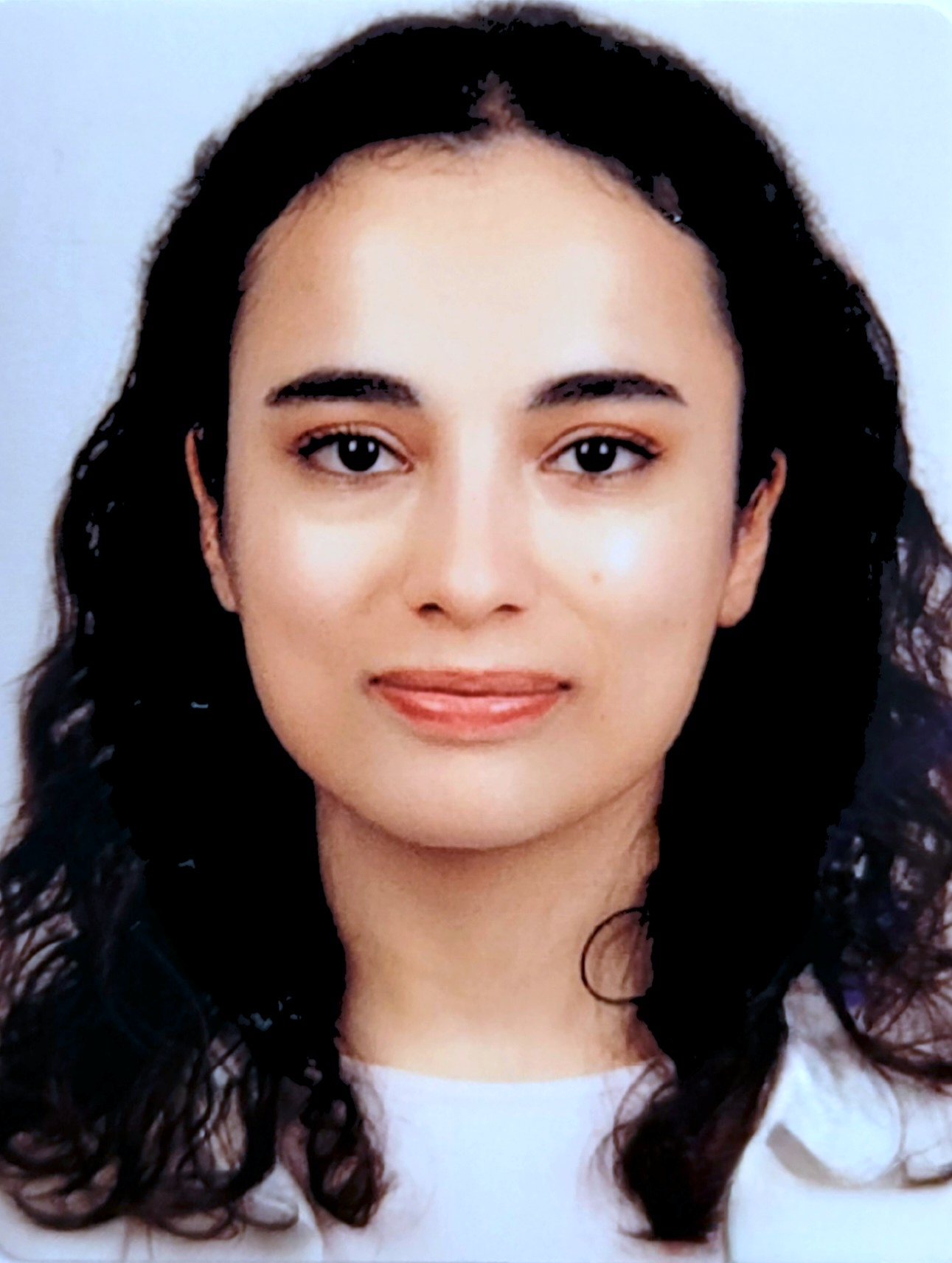}}]{Hanne Say Eren} is a doctoral
researcher at the Institute of Machine Learning and Neural Computation, Graz University of Technology, Graz, Austria. She received the BSc degree in Computer Science from TOBB University of Economics and Technology, Ankara, Turkey and MSc degree in Artificial Intelligence from \"{O}zye\u{g}in University, Istanbul, Turkey. Her research interests are lifelong learning, visual reasoning, efficient learning, and integrating human-like learning into machine learning.
\end{IEEEbiography}

\begin{IEEEbiography}[{\includegraphics[width=1in,height=1.25in,clip,keepaspectratio]{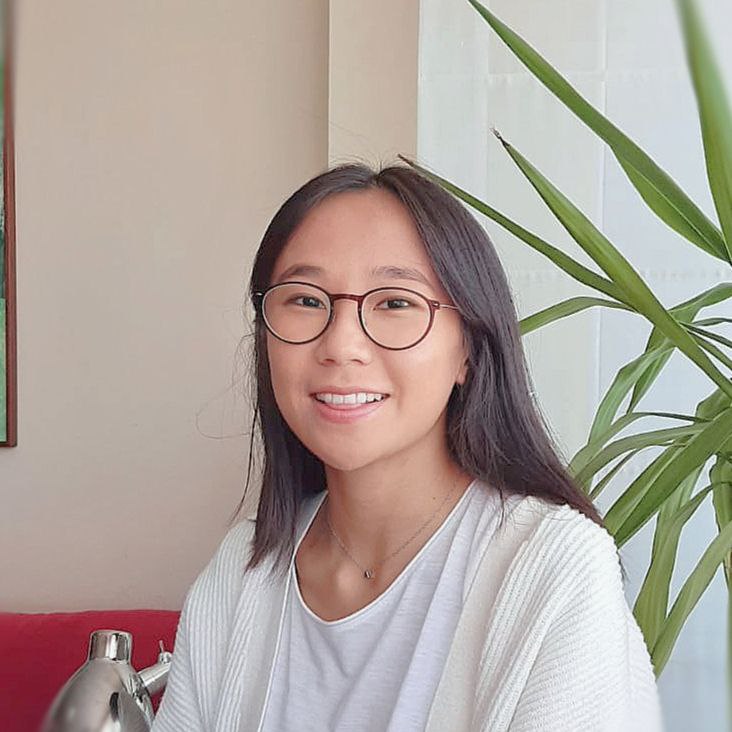}}]{Ana Ezquerro} is a doctoral researcher at the Institute of Machine Learning and Neural Computation, Graz University of Technology, Austria. She obtained her bachelor's degree in Data Science and Engineering and master's degree in Artificial Intelligence at University of A Coruña, Spain. Prior to her doctoral studies, she worked as a research technician  on tree and graph parsing. Her current research focuses on visual reasoning, graph learning, and generative models.
\end{IEEEbiography}

\begin{IEEEbiography}[{\includegraphics[width=1in,height=1.25in,clip,keepaspectratio]{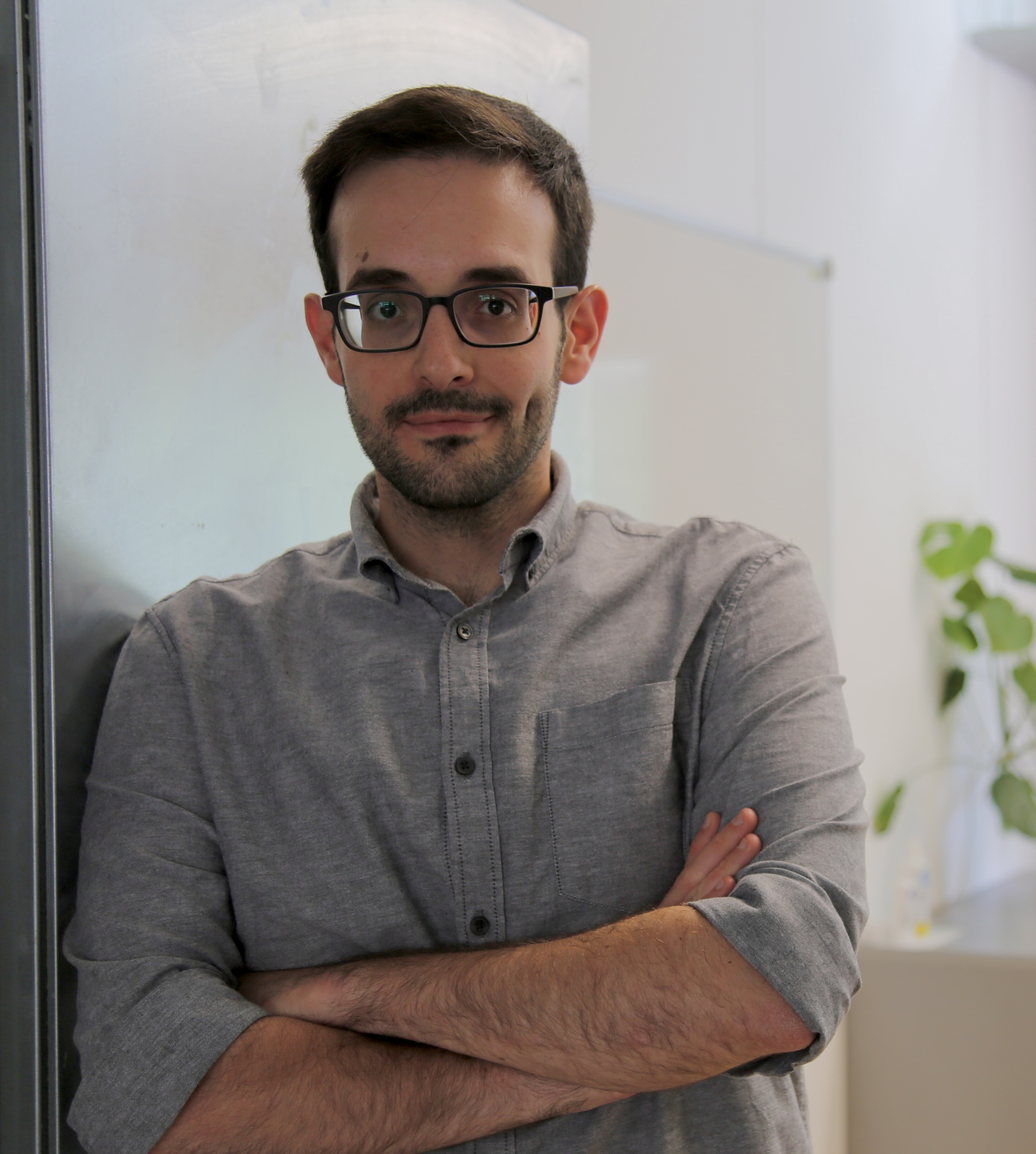}}]{Ozan \"{O}zdenizci} is an Assistant Professor at the Institute of Machine Learning and Neural Computation, Graz University of Technology, Austria. He received his PhD degree from Northeastern University, Boston, MA, USA. He previously held research appointments at the Montanuniversität Leoben, Austria, Mitsubishi Electric Research Laboratories, Cambridge, MA, USA, Max Planck Institute for Intelligent Systems, Tübingen, Germany, and Sabanci University, Istanbul, Turkey. His research focuses on developing foundational machine learning methods that enhance the reliability and efficiency of deep learning systems, with particular emphasis on robustness, security \& privacy, and resource-efficient algorithms for neural networks.
\end{IEEEbiography}

\newpage
\appendices

\setcounter{table}{0}
\setcounter{figure}{0}
\setcounter{section}{0}
\renewcommand{\thesection}{\Alph{section}}
\renewcommand{\thesubsection}{\Alph{section}\arabic{subsection}}
\renewcommand{\thefigure}{\Alph{section}\arabic{figure}}
\renewcommand{\thetable}{\Alph{section}\arabic{table}}

\section{Additional Details on Experiments}

\subsection{Hyperparameters and Model Training}\label{app:hyperparams}

We selected the hyperparameters for the reasoning tasks (see Table~\ref{tab:hyper}) using a grid search that maximized the average accuracy.
The search covered the number of DWN layers $L \in \{2, 3, 4\}$, the number of levels $\beta \in \{3, 5, 7, 9\}$, the LUT rank $n \in \{2, 4, 6\}$, the number of heads ($H$) in the $U_{rel}$ layer $\{1, 2, 4\}$ and in the subsequent DWN layers $\in \{2, 4, 8\}$, and the layer size, i.e., number of LUTs $N_U \in \{256, 512, 1024, 2048, 4096\}$.
The temperature of the popcount operation was also searched over parameter values $\lambda \in \{0.001, 0.01, 0.05, 0.07, 0.1, 0.5\}$. 
All experiments were performed on NVIDIA H100 GPUs. 

Since the model size matters in addition to accuracy for efficiency reasons, we did not simply choose the configuration with the highest validation accuracy.
Instead, we selected the model with the best trade-off between accuracy and the number of parameters, preferring the smaller configuration whenever the accuracies were comparable.

The pool size $p$ was kept as small as possible, since it determines the size of the learnable mapping matrix $W$.
The size of the first LUT layer, $U_{rel}$, is fixed for all tasks since slot dimension and threshold bits are same for all tasks ($D=64$ and $\beta=5$) which produces $320$ bits per binary slot pair $B_{ij}$.
The input to the next layer is the concatenation over all $M$ slot pairs, which has dimension $M \times 320$.
We chose $p$ empirically for each task according to this input dimension, using the smallest value that did not degrade test accuracy.

\begin{table}[h]
    \centering
    \caption{Hyperparameters of the MH-DWN blocks.}
    \label{tab:hyper}
    \resizebox{\columnwidth}{!}{%
    {\renewcommand{\arraystretch}{1.1}%
    \begin{tabular}{lccccccccc}
         \toprule
         Task & $\beta$ & $L$ & $H$ & $n$ & $N_U$ & $p$ & $\lambda$ \\
         \midrule
         \textsc{SD \& O3} & 5 & 2 & $[1, 2]$ & $[2 ,4]$ & $[320, 256]$ & 160 & 0.05 \\ 
         \textsc{RMTS} & 5 & 2 & $[1, 2]$ & $[2 ,4]$ & $[320, 512]$ & 240 & 0.05 \\
         \textsc{ID} & 5 & 2 & $[1, 2]$ & $[2 ,4]$ & $[320, 1024]$ & 160 & 0.1\\
         \textsc{Dist-3} & 5 & 2 & $[1, 2]$ & $[2 ,4]$ & $[320, 1024]$ & 160 & 0.05 \\
         \textsc{RM} & 5 & 2 & $[2, 4]$ & $[2 ,4]$ & $[320, 2048]$ & 100 & 0.07 \\
         \bottomrule
    \end{tabular}}}
\end{table}

\subsection{Details on the \textsc{RM} Task}\label{app:rmtrain}

\begin{figure}[h]
    \begin{subfigure}[c]{\columnwidth}
        \centering
        \includegraphics[scale=0.3]{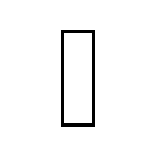}
        \hspace{0.2cm}
        \includegraphics[scale=0.3]{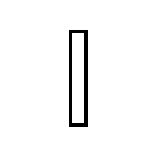}
        \hspace{0.2cm}
        \includegraphics[scale=0.3]{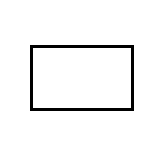}
        \hspace{0.2cm}
        \includegraphics[scale=0.3]{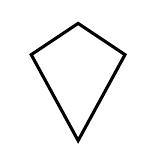}
        \hspace{0.2cm}
        \includegraphics[scale=0.3]{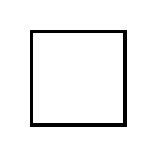}
        \vfill
        \includegraphics[scale=0.3]{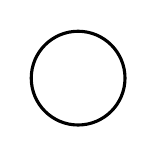}
        \hspace{0.2cm}
        \includegraphics[scale=0.3]{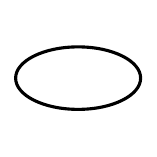}
        \hspace{0.2cm}
        \includegraphics[scale=0.3]{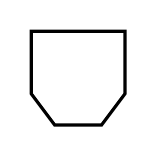}
        \hspace{0.2cm}
        \includegraphics[scale=0.3]{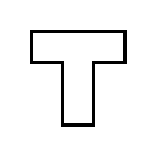}
        \hspace{0.2cm}
        \includegraphics[scale=0.3]{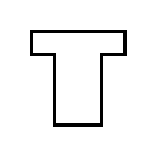}
        \vfill
        \includegraphics[scale=0.3]{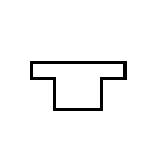}
        \hspace{0.2cm}
        \includegraphics[scale=0.3]{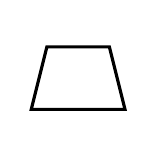}
        \hspace{0.2cm}
        \includegraphics[scale=0.3]{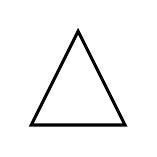}
        \hspace{0.2cm}
        \includegraphics[scale=0.3]{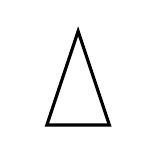}
        \hspace{0.2cm}
        \includegraphics[scale=0.3]{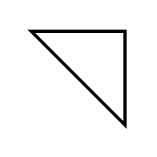}
        \caption{G-set objects.}
        \label{fig:gset}
        \vspace{0.2cm}
    \end{subfigure}
    \begin{subfigure}[c]{\columnwidth}
        \centering
        \includegraphics[scale=0.3]{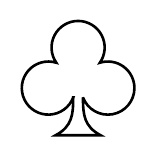}
        \hspace{0.2cm}
        \includegraphics[scale=0.3]{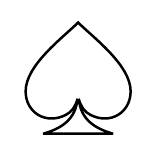}
        \hspace{0.2cm}
        \includegraphics[scale=0.3]{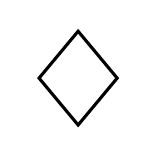}
        \hspace{0.2cm}
        \includegraphics[scale=0.15]{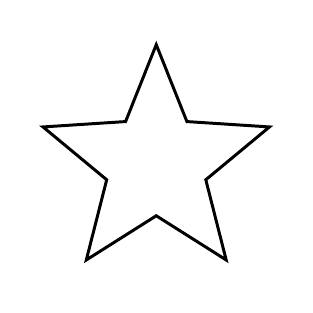}
        \hspace{0.2cm}
        \includegraphics[scale=0.15]{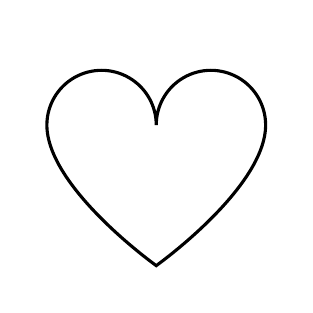}
        \caption{S-set objects.}
        \label{fig:sset}
    \end{subfigure}
    \caption{Objects used in \textsc{RM}. G-set is used for in-domain tests, and S-set is used for out-domain generalization tests.}
\end{figure}

For the \textsc{RM} tasks, to test the transferability of learned rules to unseen objects, we separate the objects into two categories, similar to DeepIQ~\cite{mandziuk2019deepiq}.
For training, we only use \textbf{G-set} objects (see Fig.~\ref{fig:gset}), which consist of simple geometric shapes such as rectangle, triangle, ellipse.
For the \textbf{S-set}, we use a different set of objects (see Fig.~\ref{fig:sset}) which are visually distinct from the G-set objects, to measure the transfer capability of models.
Other attributes used for relationships, i.e., scale, grayscale, and rotation are chosen from a discrete set of values, and are listed in Table~\ref{tab:rm-attr}.

\begin{table}[h]
    \centering
    \caption{Range of values used for object attributes in \textsc{RM}.}
    \label{tab:rm-attr}
    \begin{tabular}{lc}
    \toprule
        Attribute & Values \\
        \midrule
        Scale & [0.6, 0.8, 1] \\
        Grayscale & [0, 72, 109, 182, 255] \\
        Rotation & [0, 45, 90, 135, 180, 225, 270, 315] \\
    \bottomrule
    \end{tabular}
\end{table}

\textbf{Model training:}
All models are trained with 5K samples for each \textsc{RM} task, consisting of G-set objects only.
We trained \textsc{Slot-Dwn} for 50 epochs on \textsc{RM-1} and 100 epochs for \textsc{RM-2} and \textsc{RM-3}.
We trained OCRA~\cite{webb2023systematic} and Slot-Abstractor~\cite{mondal2024slot} models for 50 epochs on all \textsc{RM} tasks, whereas for 100 epochs for GAMR on all \textsc{RM} tasks.
For testing, we use 0.5K samples, both for in-domain $\text{G}\rightarrow \text{G}$ and out-domain $\text{G}\rightarrow \text{S}$.

\begin{table}[h]
    \centering
    \caption{Number of training epochs for ART tasks.}
    \label{tab:epochsart}
    \resizebox{\columnwidth}{!}{%
    {\renewcommand{\arraystretch}{1.1}%
    \begin{tabular}{lcccc}
    \toprule
    Task & $(m=95)$ & $(m=85)$ & $(m=50)$ & $(m=0)$ \\
    \midrule
    \textsc{SD} & 50 & 20 & 20 & 20 \\
    \textsc{RMTS} \& \textsc{Dist-3} & 100 & 50 & 50 & 50 \\
    \textsc{ID} & 50 & 50 & 50 & 50 \\
    \bottomrule
    \end{tabular}}}
\end{table}

\textbf{Slot-Attention Pretraining for \textsc{RM}:}
We pretrained the slot attention module (see Tables~\ref{tab:slot-enc} and~\ref{tab:slot-dec} for the architecture) on 150K samples in which G-set objects are randomly placed on a $3\times3$ grid. 
Training samples were resized to $128\times128$.
We also applied online augmentations, i.e., random horizontal and vertical flips (each applied independently with probability $0.5$), and a random rotation drawn uniformly from $\{0^\circ, 90^\circ, 180^\circ, 270^\circ\}$.
We trained for 500K iterations with a batch size of $64$, using the Adam optimizer with a peak learning rate of $3\times10^{-4}$.
The learning rate was linearly warmed up over the first 15K iterations, held constant for the next 5K iterations, and then decayed exponentially over 100K iterations.
We clipped the gradient norm at a maximum of $1.0$.
To accelerate the slot attention pretraining, we used PyTorch mixed-precision training.

\begin{table}[t]
\centering
\caption{Convolutional encoder architecture for \textsc{RM}.}
\label{tab:slot-enc}
\resizebox{\columnwidth}{!}{%
{\renewcommand{\arraystretch}{1.1}
\begin{tabular}{lccccc}
     \toprule
     \textsc{Type} & \textsc{Channels} & \textsc{Activation} & \textsc{Kernel} & \textsc{Stride} & \textsc{Padding} \\
     \midrule
     \textsc{2D conv} & 64 & \textsc{ReLU} & $5\times5$ & 2 & 2 \\
     \textsc{2D conv} & 64 & \textsc{ReLU} & $5\times5$ & 1 & 2 \\
     \textsc{2D conv} & 64 & \textsc{ReLU} & $5\times5$ & 1 & 2 \\
     \textsc{2D conv} & 64 & \textsc{ReLU} & $5\times5$ & 1 & 2 \\
     \textsc{Position Emb.} & - & - & - & - & - \\
     \textsc{Flatten} & - & - & - & - & - \\
     \textsc{Layer Norm} & - & - & - & - & - \\
     \textsc{Linear} & 64 & \textsc{ReLU} & - & - & - \\
     \textsc{Linear} & 64 & - & - & - & - \\
    \bottomrule
\end{tabular}}}
\end{table}

\begin{table}[t]
\centering
\caption{Spatial broadcast decoder hyperparameters for \textsc{RM}. We use an $8\times8$ grid for the positional embeddings in the spatial decoder~\cite{watters2019spatial}, instead of the latent height and width dimensions ($H=64\times W=64$), to create a bottleneck for better slot disentanglement during decoding.}
\label{tab:slot-dec}
\resizebox{\columnwidth}{!}{%
{\renewcommand{\arraystretch}{1.1}
\begin{tabular}{lccccc}
     \toprule
     \textsc{Type} & \textsc{Channels} & \textsc{Activation} & \textsc{Kernel} & \textsc{Stride} & \textsc{Padding} \\
     \midrule
     \textsc{Position Emb.} & - & - & - & - & - \\
     \textsc{2D conv} & 64 & \textsc{ReLU} & $5\times5$ & 2 & 2 \\
     \textsc{2D conv} & 64 & \textsc{ReLU} & $5\times5$ & 2 & 2 \\
     \textsc{2D conv} & 64 & \textsc{ReLU} & $5\times5$ & 2 & 2 \\
     \textsc{2D conv} & 64 & \textsc{ReLU} & $5\times5$ & 2 & 2 \\
     \textsc{2D conv} & 2 & - & $3\times3$ & 1 & 1 \\
    \bottomrule
\end{tabular}}}
\end{table}

\setcounter{table}{0}
\setcounter{figure}{0}

\section{Additional Details on Methodology}

\subsection{Post-Processing of Slot Embeddings}\label{app:postproc}

\begin{figure*}[t]
    \centering
    \begin{subfigure}[c]{\linewidth}
        \centering
        \includegraphics[scale=0.2]{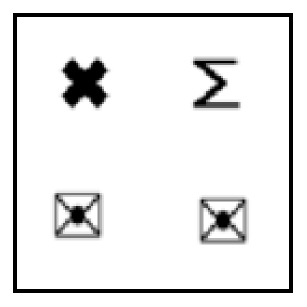}
        \hfill
        \includegraphics[scale=0.2]{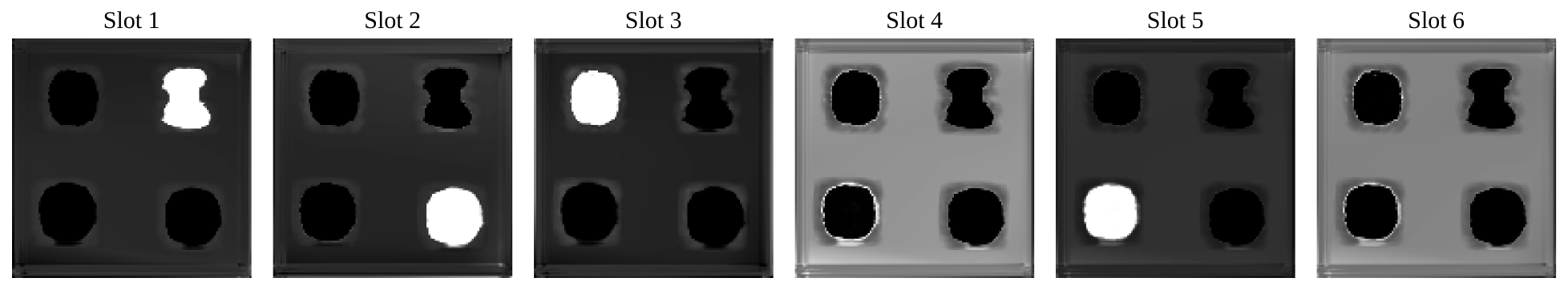}
        \hfill
        \includegraphics[scale=0.2]{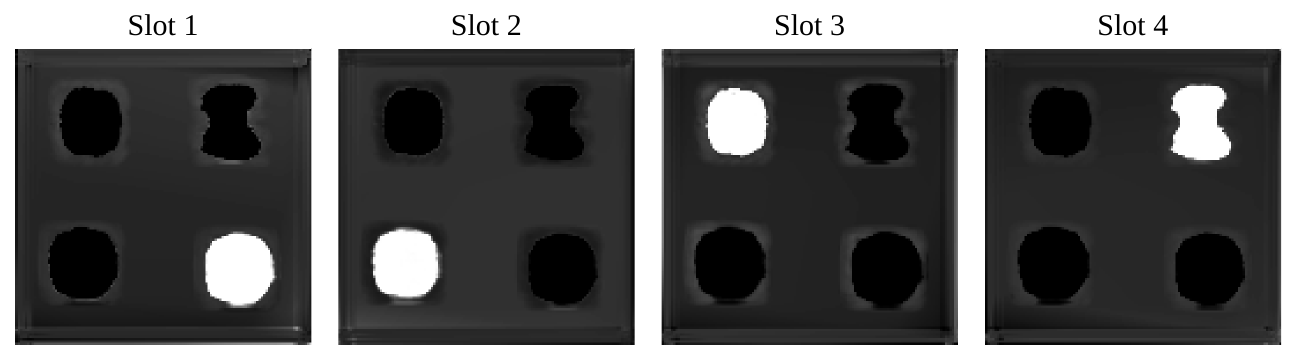}
        \hfill
        \includegraphics[scale=0.2]{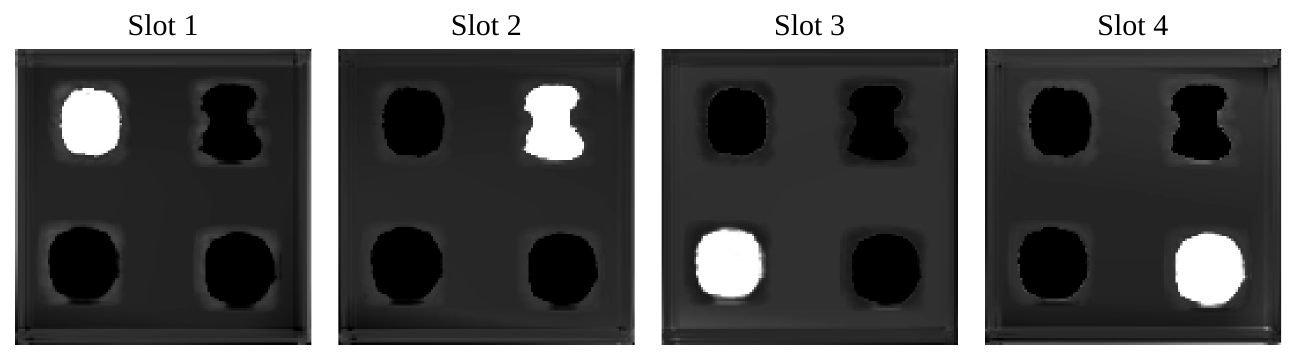}
        
        \vfill
        \includegraphics[scale=0.2]{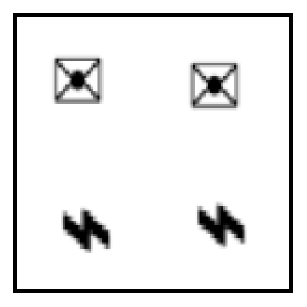}
        \hfill
        \includegraphics[scale=0.2]{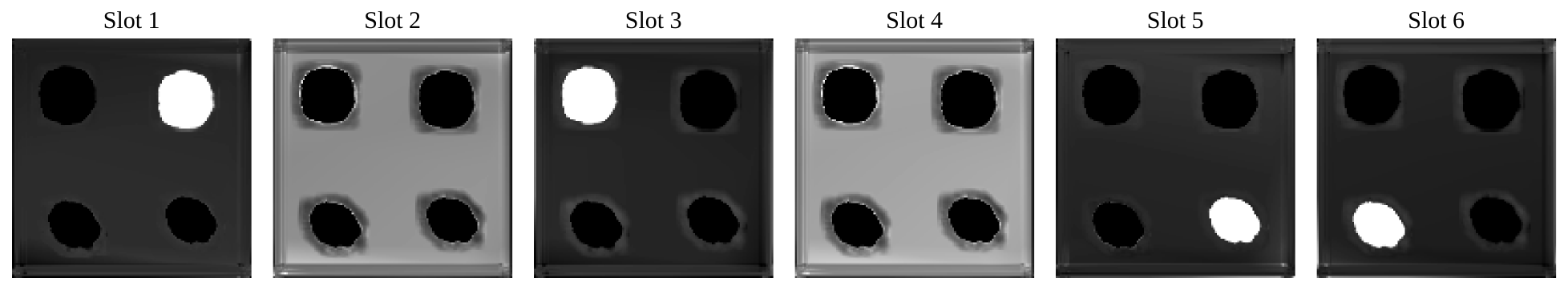}
        \hfill
        \includegraphics[scale=0.2]{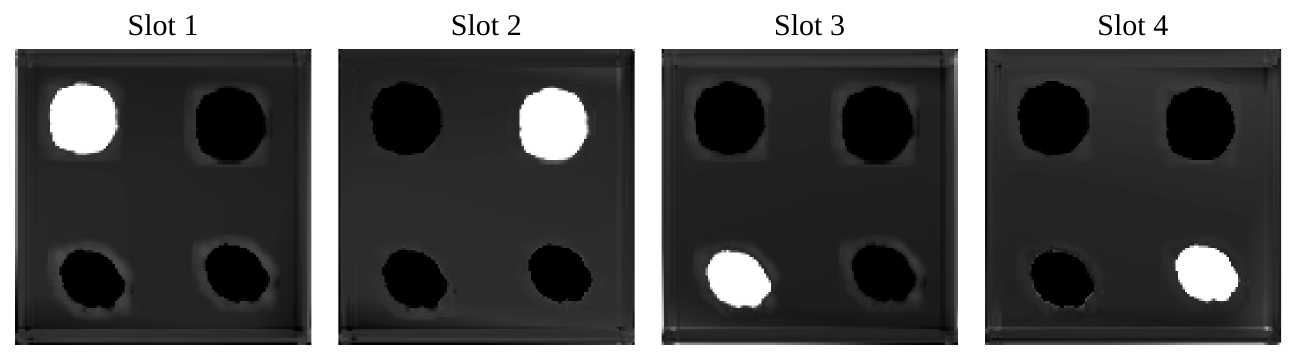}
        \hfill
        \includegraphics[scale=0.2]{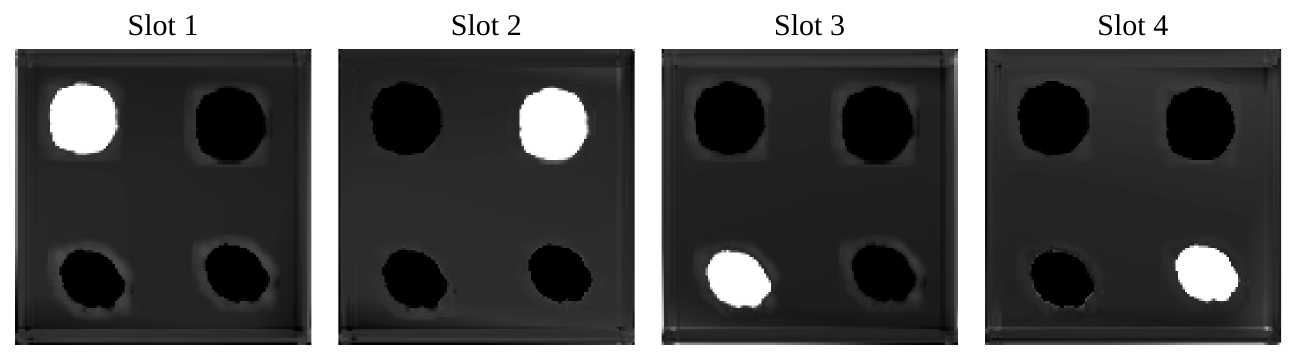}
        
        \vfill
        \includegraphics[scale=0.2]{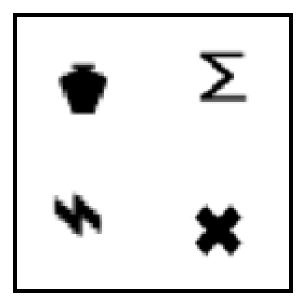}
        \hfill
        \includegraphics[scale=0.2]{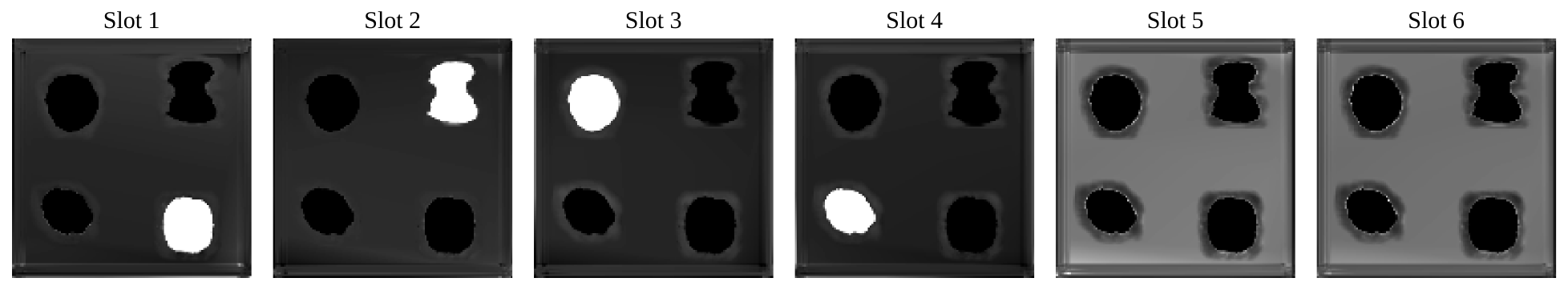}
        \hfill
        \includegraphics[scale=0.2]{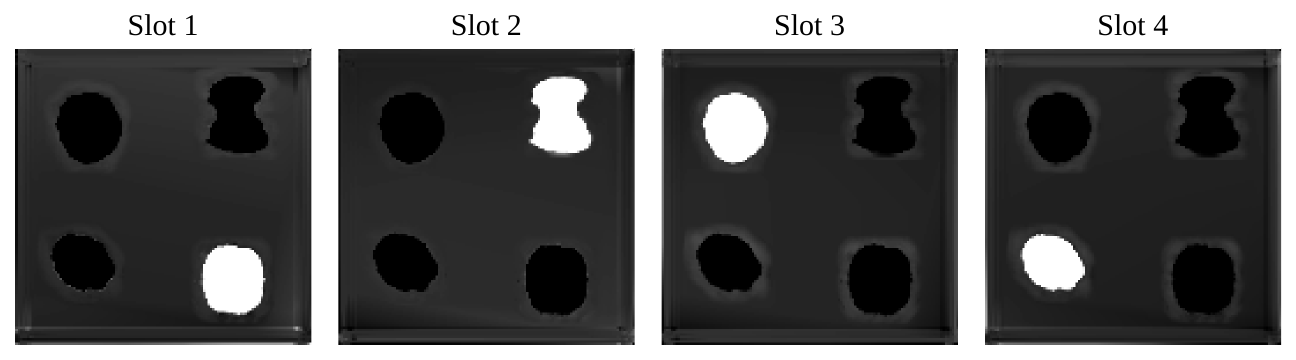}
        \hfill
        \includegraphics[scale=0.2]{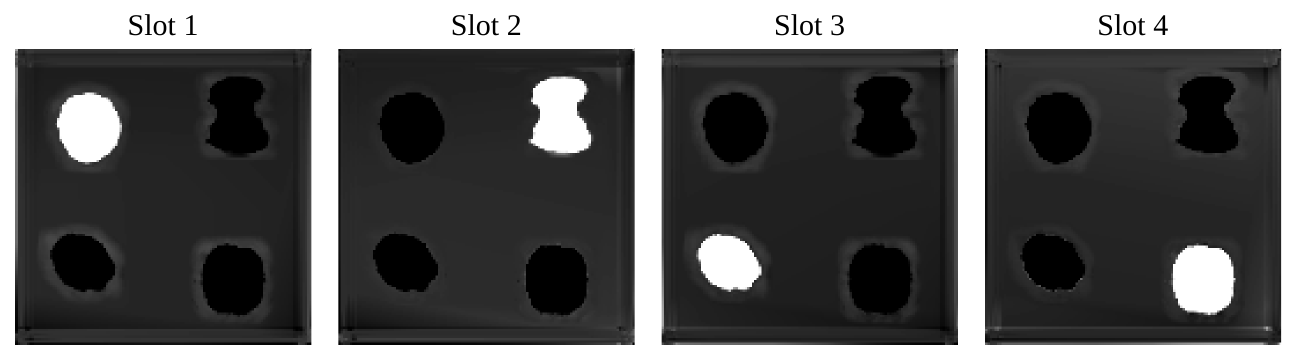}
        \caption{Original image (\textbf{left}), attention maps $\overline{attn^\tau}$ of each slot (\textbf{middle-left}), filtered slots (\textbf{middle-right}), raster-ordered slots (\textbf{right}).}
        \label{fig:rmts-postproc}
        \vspace{0.5cm}
    \end{subfigure}
    \begin{subfigure}[c]{\linewidth}
        \centering
        \includegraphics[scale=0.2]{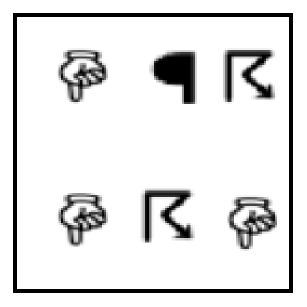}
        \hspace{0.5cm}
        \includegraphics[scale=0.2]{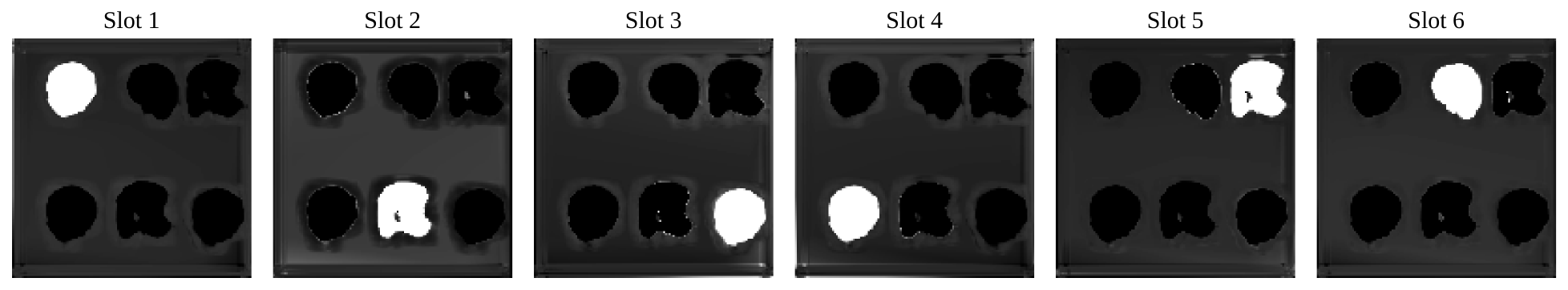}
        \hspace{0.5cm}
        \includegraphics[scale=0.2]{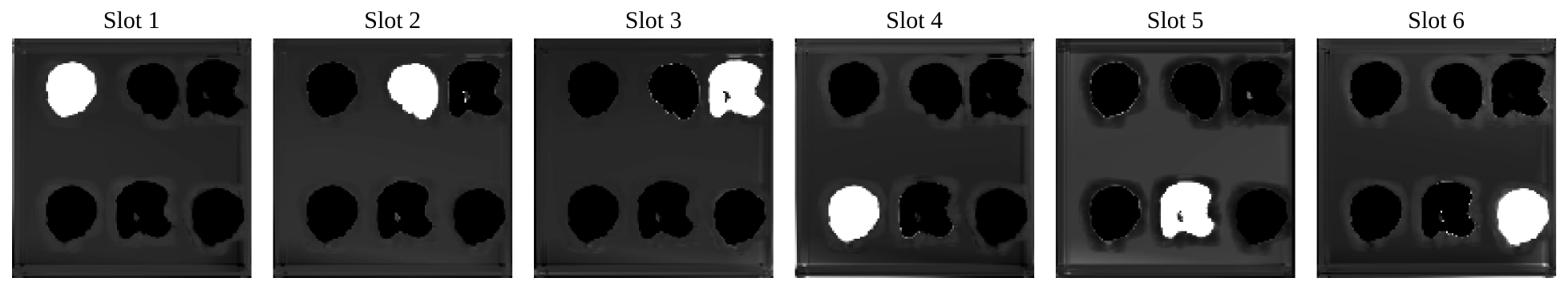}
        \vfill
        
        \includegraphics[scale=0.2]{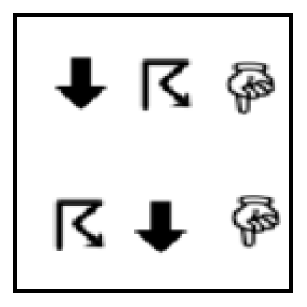}
        \hspace{0.5cm}
        \includegraphics[scale=0.2]{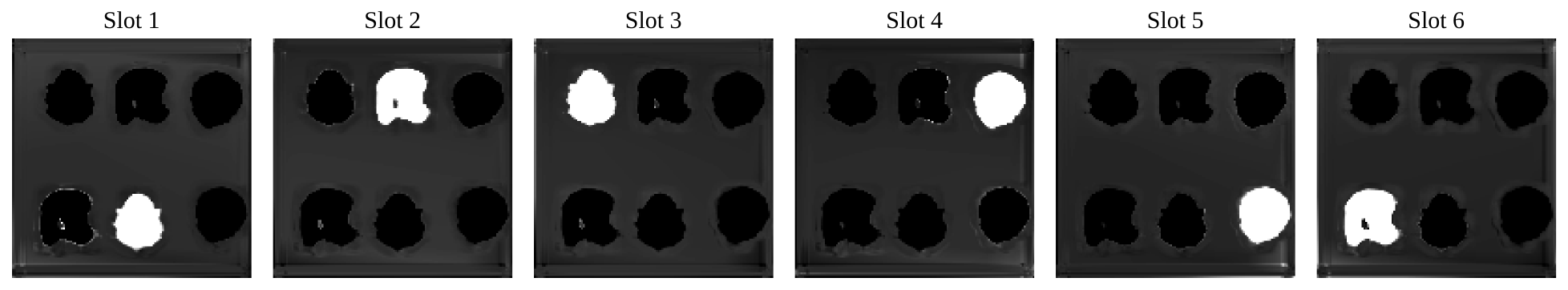}
        \hspace{0.5cm}
        \includegraphics[scale=0.2]{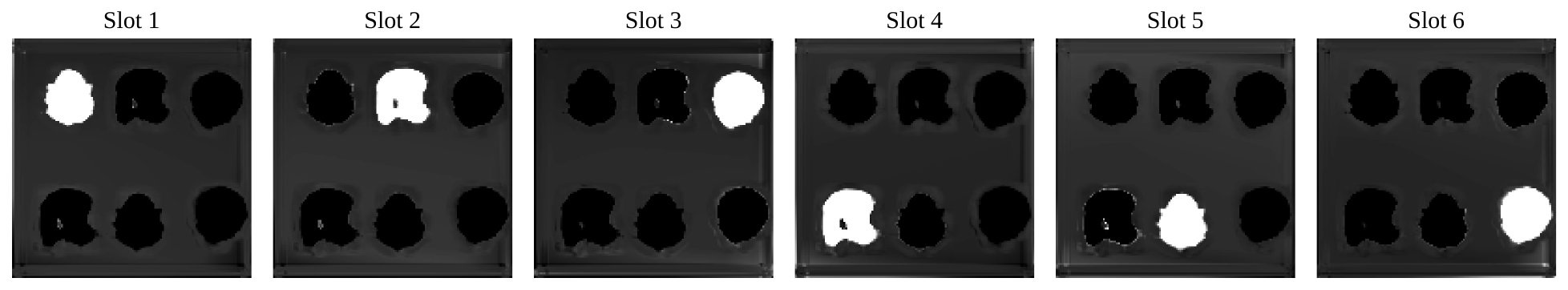}
        \vfill
        
        \includegraphics[scale=0.2]{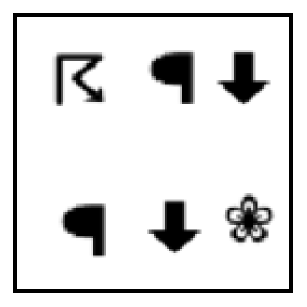}
        \hspace{0.5cm}
        \includegraphics[scale=0.2]{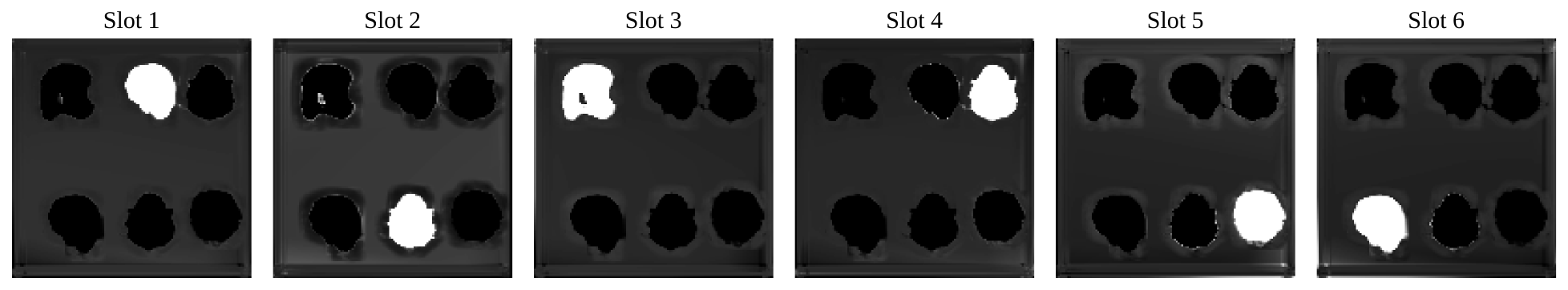}
        \hspace{0.5cm}
        \includegraphics[scale=0.2]{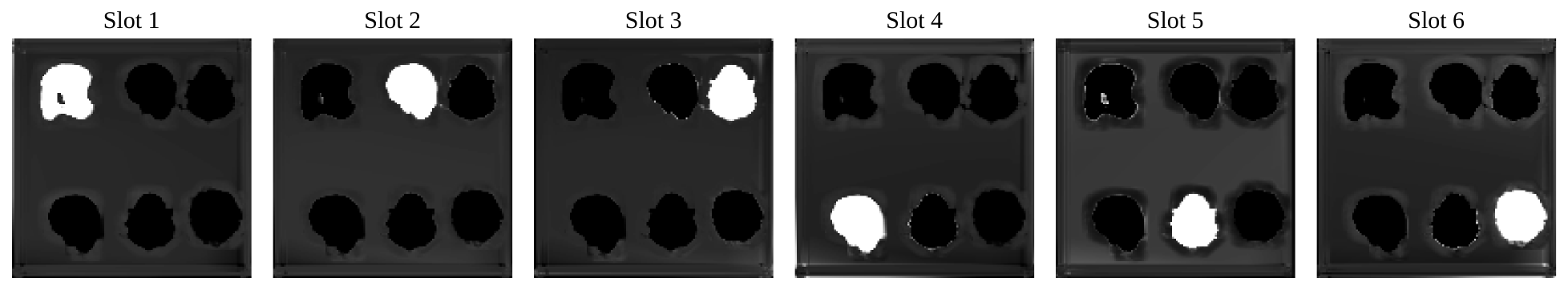}
        
        \caption{Original image (\textbf{left}), attention maps $\overline{attn^\tau}$ of each slot (\textbf{middle}), raster-ordered slots (\textbf{right}).}
        \label{fig:dist3-postproc}
    \end{subfigure}
    \caption{Demonstration of post-processing of slots on (a) \textsc{RMTS} and (b) \textsc{Dist-3}.}
    \label{fig:art-postproc}
\end{figure*}

In Fig.~\ref{fig:art-postproc}, we show the post-processing steps in \textsc{RMTS} and \textsc{Dist-3} examples. After forwarding the original panel from the pretrained slot attention, we obtain the attention maps of each slot. Then, redundant slots are filtered based on aggregated attention activity, which is followed by raster-order sorting, i.e., top-to-bottom, left-to-right.

\subsection{Sparse Learnable Mapping}\label{app:sparselm}

\begin{figure}[h]
    \centering
    \includegraphics[width=.9\columnwidth]{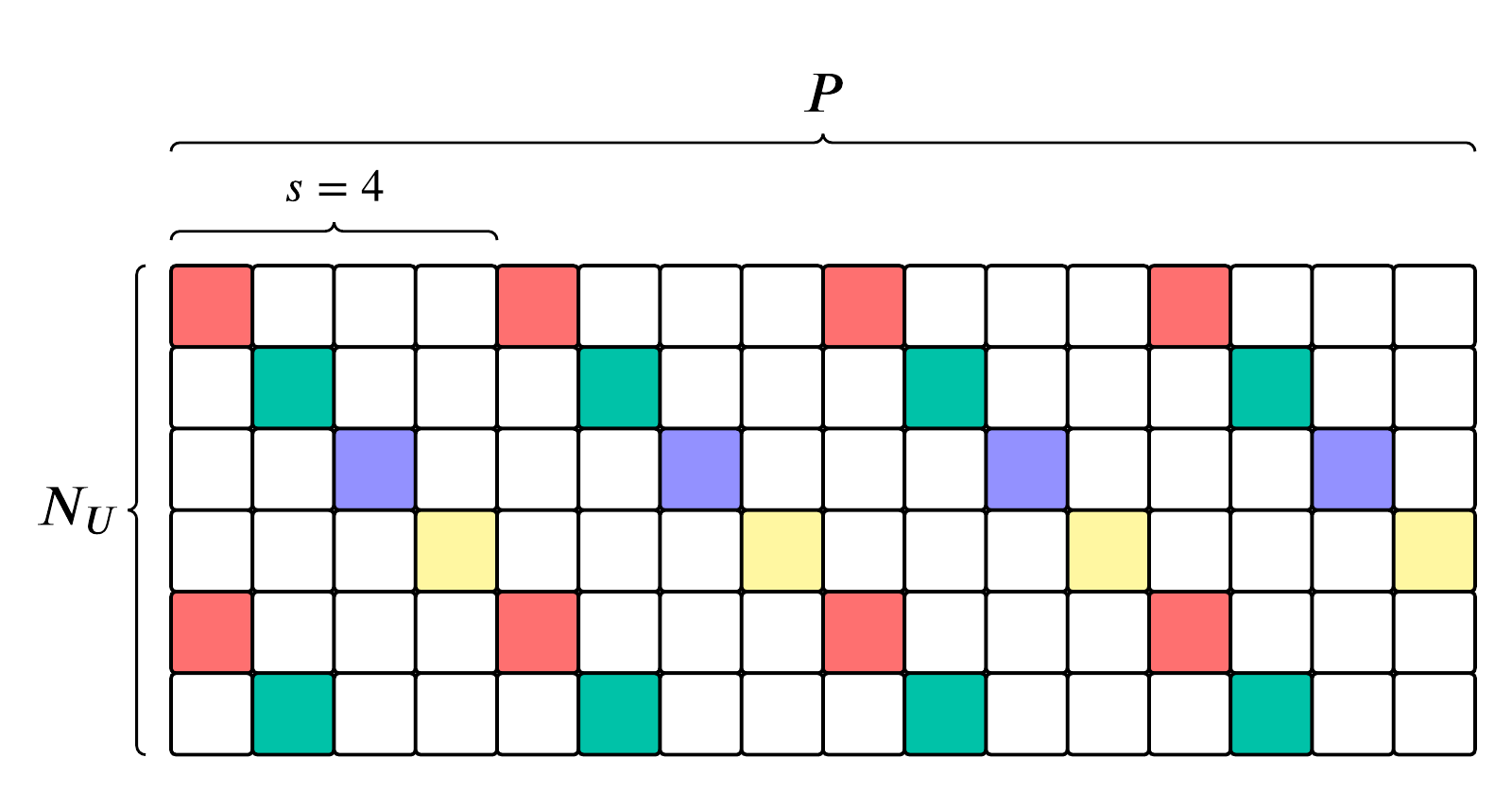}
    \caption{An illustration of strided pooling applied to the learnable matrix $W$, to create a sparse learnable mapping. We reduce the number of learnable parameters required for $W$ from $P\times n \times N_U$ to $p\times n \times N_U$.}
    \label{fig:poolstride}
\end{figure}

We changed the original learnable mapping structure of DWNs~\cite{bacellar2024differentiable} to achieve a larger reduction in the number of parameters in our model for better efficiency.
Instead of keeping the full learnable mapping matrix $W \in \mathbb{R}^{P\times n\times N_U}$, we limit the number of input bits each LUT can access to $p$ bits, where $p\ll P$.
We choose these $p$ bits in a strided manner, such that for each LUT indexed by $i \in \{0, 1, \dots N_U-1\}$ we create a pool, i.e., $\mathrm{pool}_i$, as follows: 
\begin{equation}
    \mathrm{pool}_i = \{\, j \in \{0,\dots,P-1\} \;:\; j \equiv i \!\!\pmod{s} \,\},
\end{equation}
where $s = \lfloor P/p\rfloor$ is the stride.
A visualization of strided pooling is shown in Fig.~\ref{fig:poolstride}, where rows correspond to individual LUTs, and columns correspond to input bits.
Each LUT (row) with rank $n$ is free to choose which $n$ bits to wire from the corresponding pool, shown in colors. 

Note that the strided construction gives only $s$ distinct pools, and after that the pools repeat.
We therefore require $s\leq N_U$, equivalently $p\geq P/N_U$, such that every pool is assigned to at least one LUT and no input is excluded from selection.
LUTs sharing a pool remain functionally distinct, as each retains an independent scoring over the same input subspace and an independent look-up table.

\begin{figure*}[t]
    \centering
    \begin{subfigure}[t]{\linewidth}
        \centering
        \includegraphics[width=\linewidth]{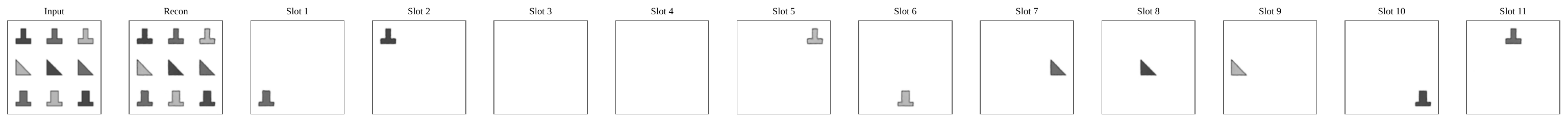}
        \includegraphics[width=\linewidth]{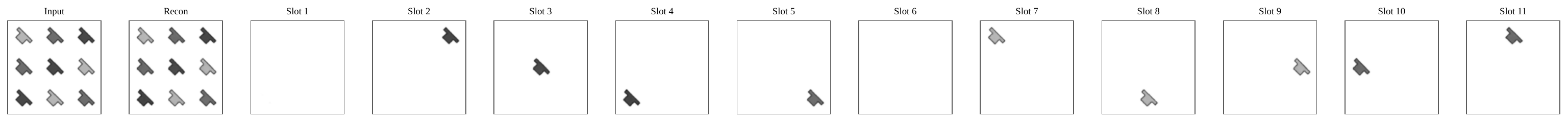}
        \includegraphics[width=\linewidth]{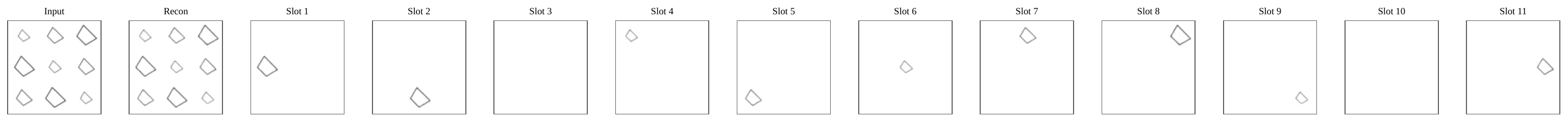}
        \includegraphics[width=\linewidth]{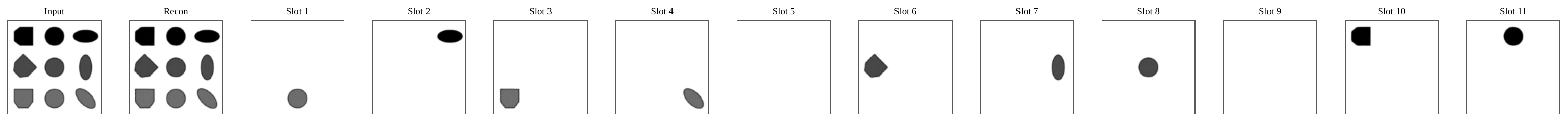}
        \includegraphics[width=\linewidth]{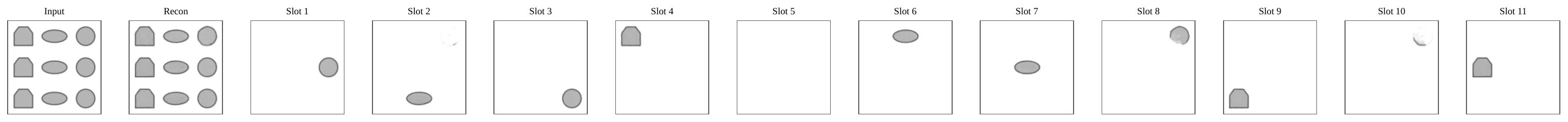}
        \includegraphics[width=\linewidth]{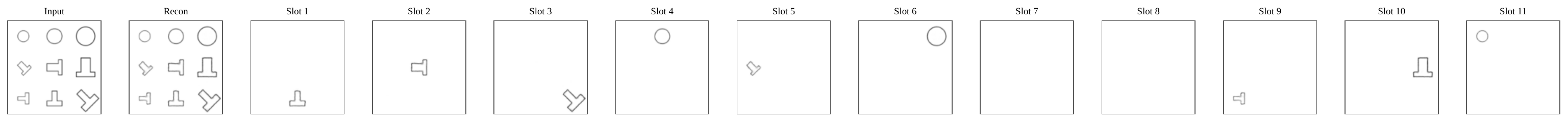}
        \includegraphics[width=\linewidth]{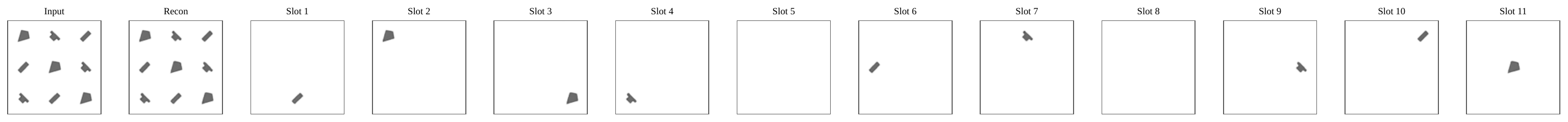}
        \caption{G-set.}
        \label{fig:gsetrecons}
    \end{subfigure}
    \begin{subfigure}[t]{\linewidth}
        \centering
        \includegraphics[width=\linewidth]{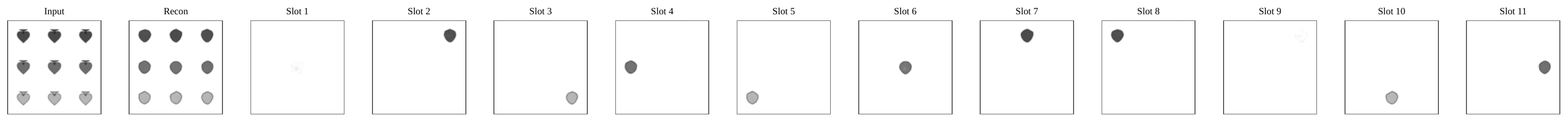}
        \includegraphics[width=\linewidth]{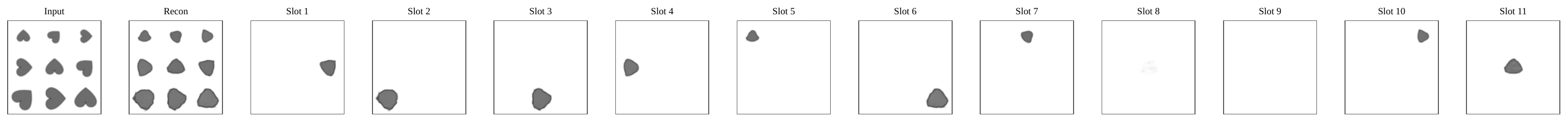}
        \includegraphics[width=\linewidth]{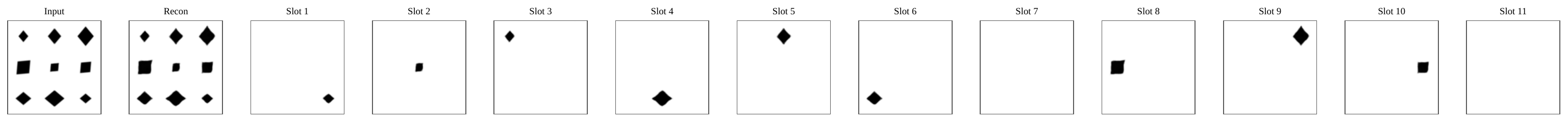}
        \includegraphics[width=\linewidth]{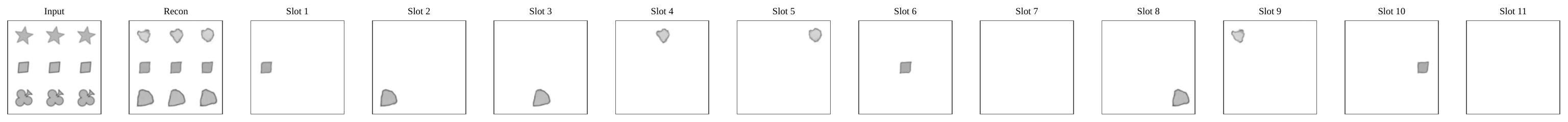}
        \includegraphics[width=\linewidth]{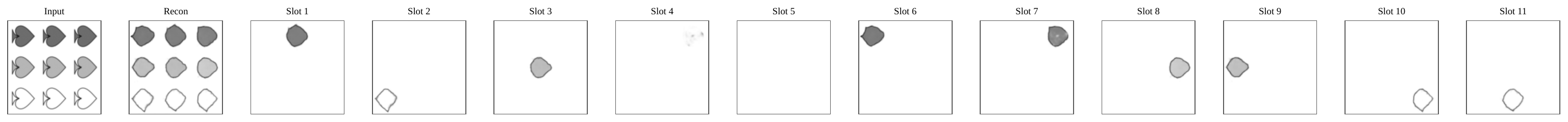}
        \includegraphics[width=\linewidth]{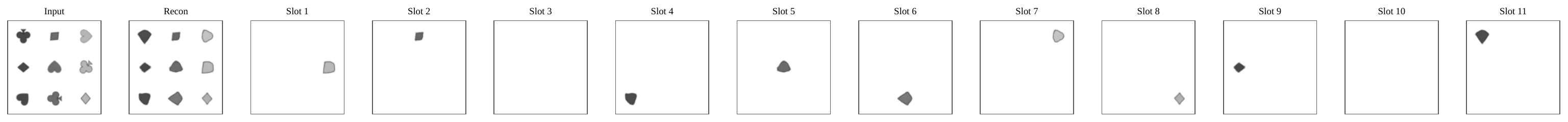}
        \includegraphics[width=\linewidth]{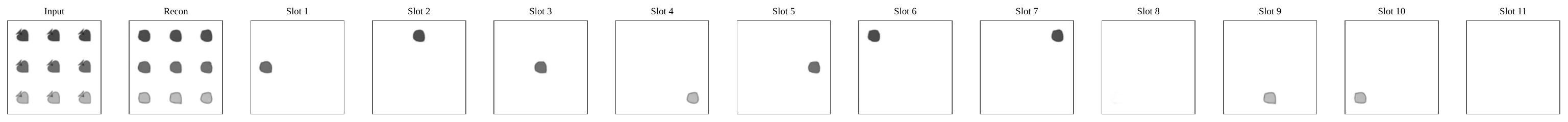}
        \caption{S-set.}
        \label{fig:ssetrecons}
    \end{subfigure}
    \caption{Visualizations of slots and input reconstructions for G-set (in domain) and S-set (out domain) samples from \textsc{RM}.}
    \label{fig:slotrecons}
\end{figure*}

\setcounter{table}{0}
\setcounter{figure}{0}

\section{Additional Experimental Results}

\subsection{Consistent Rule Learning}\label{app:sanity}

We tested whether \textsc{Slot-Dwn} is truly reasoning on the rules of the given task, by comparing the performance of the model with 2 tasks modified from the ART dataset, namely $\neg$\textsc{RMTS} and \textsc{Dist-3-Rand}.
In $\neg$\textsc{RMTS}, the task is \emph{not} matching the relation between given two objects, the opposite of \textsc{RMTS}.
The expected result would be for the model to still discover the pattern between the given objects and target objects, i.e., relation in the opposite manner, and perform similarly.
In the \textsc{Dist-3-Rand}, we randomly choose a target among the candidates and label it as the correct one, to see whether our model fails when there is no relationship between the context and the target.
This tests if our model truly reasons rather than relying on cues dictated as the target.
We show the results in Table~\ref{tab:sanity}, which confirms our model is actually capturing the underlying relational structure, instead of exploiting shortcuts.

\begin{table}[t]
    \centering
    \caption{Generalization ($m=95$) comparison of \textsc{RMTS} vs. its counterpart $\neg$\textsc{RMTS}, and \textsc{Dist-3} vs. \textsc{Dist-3-Rand}. Results show average and standard error $\left(\pm\right)$ of 10 runs.}
    \label{tab:sanity}
     \resizebox{\columnwidth}{!}{%
    {\renewcommand{\arraystretch}{1.1}%
    \begin{tabular}{lcccc}
         \toprule
           & \textsc{RMTS}  & $\neg$\textsc{RMTS}  & \textsc{Dist-3}  & \textsc{Dist-3-Rand} \\
          \cmidrule(lr){2-5}
         Accuracy $(\%)$&$88.49 \pm 0.3$ & $87.78 \pm 0.5$ & $84.43 \pm 0.3$  & $23.81 \pm 2.1$ \\
         \bottomrule
    \end{tabular}}}
\end{table}

\begin{table}[h]
    \centering
    \caption{Task accuracy vs. number of heads $H_1$ in the first LUT layer ($U_{rel}$), and $H_2$ in the second LUT layer ($U_1$).}
    \label{tab:mh-abl}
    \resizebox{\linewidth}{!}{%
    {\renewcommand{\arraystretch}{1.1}%
    \begin{tabular}{cccccccc}
    \toprule
     \multicolumn{2}{c}{\textit{heads}} & \multicolumn{2}{c}{RM-1} & \multicolumn{2}{c}{RM-2} & \multicolumn{2}{c}{RM-3} \\
     \cmidrule(lr){1-2} \cmidrule(lr){3-4} \cmidrule(lr){5-6} \cmidrule(lr){7-8}
     $H_1$ & $H_2$ & $\text{G}\rightarrow\text{G}$ & $\text{G}\rightarrow\text{S}$ & $\text{G}\rightarrow\text{G}$ & $\text{G}\rightarrow\text{S}$ & $\text{G}\rightarrow\text{G}$ & $\text{G}\rightarrow\text{S}$ \\
     \midrule
     1 & 1 & 90.34 & 78.11 & 79.89 & 67.28 & 73.51 & 63.20  \\
     1 & 2 & 92.61 & 79.44 & 81.53 & 69.47 & 75.86 & 67.14 \\
     2 & 2 & 93.94 & 81.32 & 84.18 & 73.50 & 79.33 & 69.92 \\
     \midrule
     \textbf{2} & \textbf{4} & 94.42 & 83.81 & 84.72 & 75.94 & 80.27 & 70.35 \\
     \midrule
     4 & 4 & 91.15 & 80.36 & 82.27 & 71.61 & 77.14 & 68.64 \\
     6 & 6 & 84.17 & 76.28 & 77.47 & 67.20 & 69.53 & 61.24 \\
     \bottomrule
    \end{tabular}}}
\end{table}

\subsection{Ablation Study on the Number of Heads}\label{app:mhdwn}

We further analyze the effect of the number of heads in the LUT layers of \textsc{Slot-Dwn}.
Especially, we investigate whether the proposed multi-head version MH-DWN is necessary for stronger in-domain reasoning and out-domain generalization.
We performed this analysis on \textsc{RM} since we integrated more heads for this task (see Table~\ref{tab:hyper}) and it requires learning multiple rules simultaneously.

Our results for varying $H$ configurations, for both the first layer $U_{rel}$ and the second layer $U_1$, are reported in Table~\ref{tab:mh-abl}. 
A single-head DWN, which corresponds to the original implementation~\cite{bacellar2024differentiable}, is consistently outperformed by other multi-head configurations, when $H$ is chosen appropriately.
The best configuration ($H_1=2, H_2=4$), which we use in our model, improves the single-head accuracy in a range of 4\% to 9\%.
Our model especially benefits from a multi-head $U_{rel}$ layer, where it leads to a substantial increase in both in-domain and out-domain accuracy.
However, as observed with $H_1=6, H_2=6$, increasing head count more does not necessarily lead to better performance, since this resulted in worse performance than the single-head model.

\end{document}